%% file: main.tex
\documentclass[10pt,twocolumn,letterpaper]{article}

\usepackage[pagenumbers]{cvpr} 

\input{preamble}

\definecolor{cvprblue}{rgb}{0.21,0.49,0.74}
\usepackage[pagebackref,breaklinks,colorlinks,allcolors=cvprblue]{hyperref}

\def\paperID{7844} 
\def\confName{CVPR}
\def\confYear{2026}

\title{RevoNAD: Reflective Evolutionary Exploration for Neural Architecture Design\vspace{-0.5em}}

\author{
Gyusam Chang$^1$\quad
Jeongyoon Yoon$^1$\quad
Shin han yi$^1$\quad
JaeHyeok Lee$^1$\quad
Sujin Jang$^2$\quad
Sangpil Kim$^1$\thanks{Corresponding authors.} \\
$^1$Korea University \qquad $^2$Samsung AI Center\\
{\tt\small $\{$gsjang95, dabujin98, hannie12, dlwogurgur, spk7$\}$@korea.ac.kr} \quad
{\tt\small s.steve.jang@samsung.com} \\
\vspace{-1em}}

\begin{document}

\maketitle
\footnotetext[2]{Project page: \url{https://kuai-lab.github.io/RevoNAD}}
\input{sec/0_abstract}

\input{sec/1_intro}
\input{sec/2_related}

\input{sec/3_method}

\input{sec/4_exp}

\input{sec/5_con}

{
    \small
    \bibliographystyle{ieeenat_fullname}
    \bibliography{main}
}

\input{sec/X_suppl}

\end{document}

%% file: preamble.tex
\usepackage{natbib}
\usepackage{multirow}
\usepackage{booktabs}
\usepackage{graphicx}
\usepackage{colortbl}
\usepackage{amsmath}
\usepackage{bm}
\usepackage{diagbox}
\usepackage{tabularx}
\usepackage{caption}
\usepackage{subcaption}
\usepackage{amssymb}

\usepackage{fancyhdr}
\usepackage[utf8]{inputenc} 
\usepackage{url}            
\usepackage{amsfonts}       
\usepackage{nicefrac}       
\usepackage{microtype}      
\usepackage{pifont}
\usepackage{array}

\usepackage[boxed, ruled, linesnumbered]{algorithm2e}
\DontPrintSemicolon
\SetKwComment{tcp}{\# }{}

\usepackage{minitoc}

\usepackage{tikz}
\usetikzlibrary{arrows.meta, positioning, shapes.geometric}

\usetikzlibrary{fit, calc}
\usepackage{pgfplots}
\pgfplotsset{compat=1.18}

\usepackage{subcaption}

\newcommand{\coloredrowcell}[2]{\rowcolor[HTML]{#1}\cellcolor[HTML]{FFFFFF} #2}
\newcommand{\coloredcell}[2]{\cellcolor[HTML]{#1} #2}

\newcolumntype{C}[1]{>{\centering\arraybackslash}m{#1}}

\newcommand{\myparagraph}[1]{\vspace{2pt}\noindent{\bf #1}}

\def\ie{\emph{i.e.}}
\def\eg{\emph{e.g.}}

\AtEndPreamble{
    \usepackage[capitalize]{cleveref}
    \crefname{section}{Sec.}{Secs.}
    \Crefname{section}{Section}{Sections}
    \Crefname{table}{Table}{Tables}
    \crefname{table}{Tab.}{Tabs.}
}

\makeatletter
\newcommand{\appendixsetup}{%
  \appendix
  \setcounter{tocdepth}{2}%
  \let\origaddcontentsline\addcontentsline
  \renewcommand{\addcontentsline}[3]{%
    \def\@@file{##1}%
    \def\@@toc{toc}%
    \ifx\@@file\@@toc
      \origaddcontentsline{app}{##2}{##3}%
    \fi
  }%
}

\newcommand{\appendixtableofcontents}{%
  \begingroup
  \setcounter{tocdepth}{2}
  \renewcommand{\contentsname}{Contents}
  \section*{\contentsname}
  \@starttoc{app}
  \endgroup
  \bigskip
}
\makeatother

%% file: sec/0_abstract.tex
\begin{abstract}
Recent progress in leveraging large language models (LLMs) has enabled Neural Architecture Design (NAD) systems to generate new architecture not limited from manually predefined search space. 
Nevertheless, LLM-driven generation remains challenging: the token-level design loop is discrete and non-differentiable, preventing feedback from smoothly guiding architectural improvement.
These methods, in turn, commonly suffer from mode collapse into redundant structures or drift toward infeasible designs when constructive reasoning is not well grounded.
We introduce RevoNAD, a reflective evolutionary orchestrator that effectively bridges LLM-based reasoning with feedback-aligned architectural search. 
First, RevoNAD presents a Multi-round Multi-expert Consensus to transfer isolated design rules into meaningful architectural clues. 
Then, Adaptive Reflective Exploration adjusts the degree of exploration leveraging reward variance; it explores when feedback is uncertain and refines when stability is reached.
Finally, Pareto-guided Evolutionary Selection effectively promotes architectures that jointly optimize accuracy, efficiency, latency, confidence, and structural diversity.
Across CIFAR10, CIFAR100, ImageNet16-120, COCO-5K, and Cityscape, RevoNAD achieves state-of-the-art performance. Ablation and transfer studies further validate the effectiveness of RevoNAD in allowing practically reliable, and deployable neural architecture design.
\vspace{-1em}
\end{abstract}

%% file: sec/1_intro.tex
\section{Introduction}
\label{sec:intro}

\begin{figure}
    \centering
    \includegraphics[width=1\linewidth]{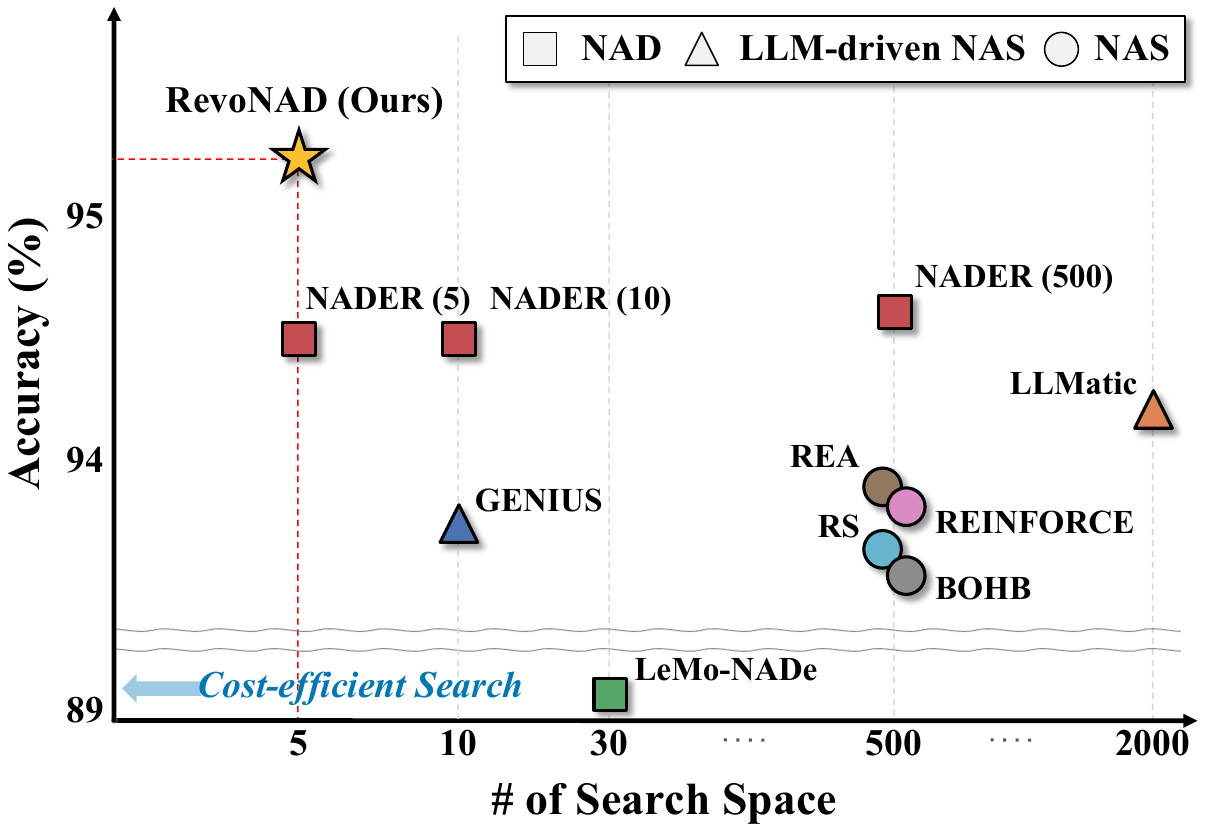}
    \vspace{-1.5em}
    \caption{Comparison of search efficiency on CIFAR10. RevoNAD finds stronger architectures with fewer trials, showing substantially improved design efficiency compared to prior NAD/NAS methods.}
    \label{fig:teaser}
    \vspace{-.5em}
\end{figure}

Neural Architecture Design (NAD) aims to synthesize high-level architectural principles into concrete design patterns and interpretable structural details. 
Unlike traditional neural architecture search (NAS) settings~\cite{chen2020drnas, xie2018snas, liu2018darts, zhang2021idarts, falkner2018bohb}, NAD does not merely select low-level hyperparameters, but instead focuses on designing the core computational topology, including hierarchical scaling, connectivity routing, feature mixing, normalization strategies, and efficiency trade-offs.
Recent progress in NAD has been driven by large language models (LLMs)~\cite{achiam2023gpt, liu2024deepseek, bai2023qwen}, which capture functional and logical relations rather than depend on predefined heuristics.

Thanks to the capacity of LLMs to encode structural patterns and theoretical insights from literature, recent NAD method~\cite{yang2025nader} has demonstrated an ability to \emph{go beyond manually predefined search spaces}.
Nevertheless, neural architecture design remains fundamentally discrete and non-differentiable regime, where training-based feedback is inherently infeasible to align with architectural generation.
As shown in~\cref{fig:teaser}, existing works navigate a highly complex and intricate architectural search space, resulting in inefficient exploration and limited accuracy gain.
Consequently, NAD frequently struggles with these systematic limitations: Its gradient-agnostic exploration often causes mode collapse, forcing the search on a small set of structurally repetitive and redundant schemata, which in turn limits both design and search efficiency.
Building on these insights, we aim to bridge LLM-based reasoning with a stable, feedback-aligned evolutionary framework that substantially encourages scalable search while preserving architectural validity.

To this end, we propose RevoNAD, a reflective evolutionary exploration orchestrator for scalable and reliable neural architecture design.
RevoNAD consists of three main modules; Multi-round Multi-expert Consensus (MMC), Adaptive Reflective Design Exploration (ARDE), and Pareto-guided Evolutionary Selection (PES). 
First, MMC decomposes fragmented architectural heuristics into conceptual sub-axes, where specialized agents collaboratively refine them into coherent and interpretable structural reasoning. 
Specifically, MMC allows multiple agents to iteratively discuss and distill local architectural clues into practical drop-in knowledge, converging toward a stable consensus and a performance-quality plateau.
The design is further guided by ARDE, which effectively adjusts exploration–exploitation intensity based on reward variance~\cite{tokic2010adaptive, rodrigues2009dynamic, dabney2020temporally}.
It adaptively explores when joint dynamics are unstable and refines promising design proposals once convergence stabilizes.
Reflective summaries then update the contextual memory that conditions subsequent prompts, significantly enhancing evolutionary reasoning based on accumulated design feedback.
As a final step, we propose PES, which prioritizes architectures that jointly optimize accuracy, efficiency, latency, confidence, and structural diversity, thereby reducing premature convergence toward overfitted models.
Subsequent generations are in turn grounded in these selected architectures, which establish well-structured design principles for reflective evolutionary development.
Taken together, RevoNAD couples structured multi-agent reasoning with adaptive exploration and Pareto-guided selection, enabling the discovery of diverse, robust, and well-informed architectures across varying compute and deployment environments.

In summary, our design orchestrator RevoNAD replaces hand-crafted NAS search spaces with an adaptively shaped design space of architectural schemata, obtained by iteratively pruning and refining designs through a reflective evolutionary loop. 
This strategy not only improves search efficiency but also yields interpretable architectural patterns that can be reliably deployed across diverse tasks. 
Extensive experiments on CIFAR10, CIFAR100, ImageNet16-120, COCO-5K, and Cityscapes, RevoNAD achieves state-of-the-art or highly competitive performance, thereby paving the way for scalable and interpretable neural architecture design.
Here, the key contributions of this work are as follows:

\begin{itemize}
\item We introduce a Multi-round Multi-expert Consensus mechanism that converts fragmented design heuristics into informative, reusable architectural inspiration tokens.

\item We propose an Adaptive Reflective Exploration that modulates LLM-driven search behavior via reward variance, balancing exploration-exploitation in a principled manner.

\item We develop a Pareto-guided Evolutionary Selection mechanism that preserves design diversity while promoting practically strong architectures across accuracy, efficiency, and broader structural robustness metrics.

\item We provide comprehensive empirical validation for NAD on CIFAR10/100, ImageNet16-120, COCO-5K, and Cityscapes, demonstrating state-of-the-art, along with detailed ablations and transfer studies.
\end{itemize}

%% file: sec/2_related.tex
\section{Related works}
\label{sec:related}

\subsection{Large Language Model and Multi-Agent}
Large Language Models (LLMs) have evolved into general purpose agents capable of planning, tool use, and self reflection ~\cite{shinn2023reflexion, zhao2024expel}. Systems such as Visual ChatGPT~\cite{wu2023visual} and HuggingGPT~\cite{shen2023hugginggpt} orchestrate multiple models under LLM control to execute complex multi step workflows ~\cite{wu2023visual, shen2023hugginggpt}.
A parallel line of work studies multi agent LLM collaboration, including role playing with specialized roles and decomposed tasks ~\cite{yu2024beyond, zhang2025roleplot, he2025crab, zhang2025omnicharacter, chen2024comm}, debate and voting schemes that independently propose and then reconcile solutions ~\cite{kaesberg2025voting, choi2025debate, feng2024m, chen2025debatecoder, park2024predict, liang2023encouraging}, and group discussion or negotiation style protocols ~\cite{zhao2024auto, liu2025dual, wang2025debt, dougru2025taking, nan2023evaluating, kim2024mdagents}. 
Recent studies highlight that adapting collaboration strategies to task complexity is increasingly crucial in high stakes settings.

\subsection{Neural Architecture Search}
Neural Architecture Search (NAS) aims to automatically find high-performing architectures within a given search space. Early methods relied on reinforcement learning, evolutionary algorithms, and Bayesian optimization ~\cite{tian2020off, liu2019dartsdifferentiablearchitecturesearch,guo2020single, nguyen2021optimal}, but incurred high computational cost due to repeated training of candidate models. To improve efficiency, weight sharing approaches train a single super network from which many sub networks are sampled and evaluated ~\cite{xie2018snas, xu2019pc, ye2022b, zhang2021idarts, williams1992simple, falkner2018bohb, dong2020bench, real2019regularized}. 
However, both classical and weight-sharing NAS methods still rely on predefined search spaces, limiting their ability to discover genuinely novel structures. 
Recent LLM-based NAS methods~\cite{zheng2023can,chen2023evoprompting,nasir2024llmatic} explore this direction by using LLMs as search operators within fixed spaces, but they remain constrained by these predefined domains in practice. 
These limitations motivate a shift from constrained search toward more diverse neural architecture design, which our work targets using LLM-based multi-agent systems.


\begin{figure*}[t]
    \centering
    \vspace{-1em}
    \includegraphics[width=1.0\linewidth]{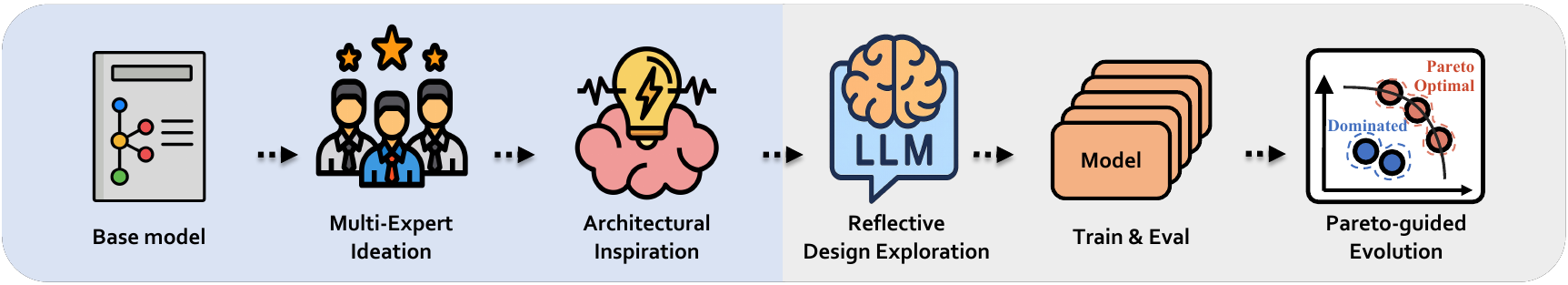}
    \vspace{-2em}
    \caption{RevoNAD Orchestrator. Starting from a base model, a multi-expert ideation module distills literature into architectural inspirations, which are then fed to an LLM-based reflective design exploration module that proposes new candidate architectures. The candidates are trained and evaluated, and a Pareto-guided evolution preserves diverse high-quality models while updating survivors for the next generation.}
    \label{fig:main}
    \vspace{-1em}
\end{figure*}

\subsection{Neural Architecture Design}
Neural Architecture Design (NAD) extends classical Neural Architecture Search by removing the constraint of a fixed, human-designed search space and instead targeting open-ended generation and refinement of architectures. Recent multi-agent LLM frameworks use role playing, debate, and voting-based collaboration to coordinate specialized agents on such complex generation and refinement tasks~\cite{du2023improving,li2023camel,liang2023encouraging,wang2022self,wu2024autogen,xi2025rise}. 
Within this line of work, NADER~\cite{yang2025nader} reframes NAD as a multi-agent collaboration problem through specialized LLM agents and a graph-based architecture representation. While this formulation improves NAD beyond predefined search spaces, its collaboration protocol and use of long-horizon experience remain largely fixed. We address these limitations by introducing adaptive collaboration and experience handling, enabling more flexible exploration and producing stronger architectures overall.

%% file: sec/3_method.tex
\section{Methods}
\label{sec:methods}


\subsection{Preliminary} 
We formulate RevoNAD, a reflective evolutionary exploration framework for NAD (see~\cref{fig:main}), as an architectural design orchestrator that bridges LLM-driven constructive reasoning with structured exploration over the architecture space $\mathcal{A}=\{a^{(n)}\}$. To this aim, we propose a memory of design cues (``\emph{inspirations}'') $\mathcal{I}_t=\{i_t^{(n)}\}$, which provides reusable structural tokens that guide architectural transformations~(\cref{sec:MMC}). 
At this stage, the exploration module proposes informative inspiration $i_t^{(n)}$ by applying transformations to successful parent $a_{t}=\mathcal{T}(a_{t-1}^{(n)}, i_t^{(n)})$, while an exploration--exploitation controller regulates redundancy and novelty~(\cref{sec:ARDE} and \cref{alg:ARDE}). 
Each candidate is trained under a uniform protocol~\cite{dong2020bench}, and we evaluate it using multiple interpretable objectives.
Finally, Pareto-based selection~(\cref{sec:PES} and \cref{alg:PES}) yields the survivor design set $\mathcal{A}_{t}$ for the next generation, while the utilities of inspirations $\mathcal{I}_t$ are updated to encourage effective architectural search, thereby completing one evolutionary cycle.


\subsection{Multi-expert Multi-round Consensus}
\label{sec:MMC}
\myparagraph{Motivation.}
LLMs are considered to be inherently versatile: \textit{they can interpret ideas, reorganize reasoning steps, and adapt to new conceptual spaces}.
Especially, this capability suggests that LLMs, in principle, serve as powerful engines for model-based design synthesis (\ie, Neural Architecture Search and Design~\cite{yang2025nader, nasir2024llmatic, liu2018darts, chen2020drnas, rahman2024lemo}).
Nevertheless, we observe that this approach typically extracts a flat set of inspirations $\mathcal{I}$ from a single perspective, failing to capture how design concepts are functionally or logically connected.
In particular, prior methods~\cite{yang2025nader, nasir2024llmatic, zheng2023can} often lack effective interactive mechanisms for refinement and verification, causing the design strategy to converge early to shallow and undifferentiated architectural clues.
As a result, the generated architectures may lead to limited diversity and weak adaptability, struggling to deal with various deployment environments, or unfamiliar task objectives.

\myparagraph{Approach.}
To tackle this limitation, we introduce a Multi-expert Multi-round Consensus (MMC), which converts isolated design rules into global structural information.
We decompose literature into distinct architecture-influential axes automatically derived from our extraction prompt. 
Each expert specializes in one conceptual dimension (\eg, hierarchical scaling, spatial–channel decoupling) and then formulates concise, drop-in architectural signals such as \texttt{DWConv→SE→MLP} or \texttt{token gating with deformable shift}. 
It is noteworthy that this decomposition can transfer individual architectural reasoning into interpretable global reasoning processes, enabling more precise and modular ideations.
Subsequently, we design a multi-round expert-consensus ideation that enables iterative refinement and convergence of individually proposed architectural ideas. 
Rather than relying on a single-shot extraction, our framework organizes structured discussions among experts to jointly encourage inspirations over multiple rounds. 
Precisely, given a paper $P$ and its sub-axes 
$\mathcal{X}=\{x_1,\dots,x_K\}$, each expert $E_k$ (associated with $x_k$) proposes a local design clue $p_k^{(t)}=E_k(P,\mathcal{I}^{(t-1)})$ at round $t$, where $\mathcal{I}^{(t-1)}$ represents the shared inspiration set from the previous round. 
The consensus module $\Omega$ then integrates these proposals into an updated set $\mathcal{I}^{(t)}=\Omega(\{p_k^{(t)}\}_{k=1}^{K})$, effectively promoting complementarity and reducing redundancy.
Here, the discussion round continues until both \textit{consensus stability} and \textit{quality plateau} are reached:
\begin{equation}
\small{\text{Jaccard}(\mathcal{I}^{(t)},\mathcal{I}^{(t-1)})\ge\tau_J,\quad
\frac{|\mu^{(t)}-\mu^{(t-1)}|}{\mu^{(t-1)}}\le\tau_Q,}
\end{equation}
where $\mu^{(t)}$ denotes the mean quality score of inspirations at round $t$, and $\tau_J$ and $\tau_Q$ specify the tolerance levels.
This adaptive termination ensures that ideation halts only when the architectural reasoning has converged, producing a stable yet diverse set of design primitives. In practice, this expert-consensus formulation leads to more constructive and informative inspirations $\mathcal{I}$ compared to a single-shot extraction, effectively bridging localized architectural signals. Note that further details are provided in Appendix~\cref{app:mmc}.

\begin{algorithm}[t]
\caption{Variance-guided ARDE}
\label{alg:ARDE}
\KwIn{Initial reflection memory $\mathcal{M}_0 = \emptyset$; 
exploration range $[\varepsilon_{\min}, \varepsilon_{\max}]$; decay $\lambda$; number of steps $N$}                        
\KwOut{Improved architecture inspirations $\{\mathbf{i}_n\}$}

\BlankLine
\textbf{Procedure:}

\For{$n = 1, 2, \dots, N$}{
  Observe current state $s_n$ (textual reflections, recent rewards, context embeddings)\;

  Compute exploration coefficient:
  \[
  \varepsilon_n = \varepsilon_{\min} + (\varepsilon_{\max}-\varepsilon_{\min})
  \cdot e^{-\lambda \cdot \text{Var}(r_{n-m:n})}
  \]

  \eIf{random() $< \varepsilon_n$}{
    \tcp{Exploration via reflective reasoning}
    $i_n \leftarrow 
    \mathcal{A}_{\text{LLM}}\!\big(\text{prompt}(s_n, \text{Reflect}(s_{n-1}, r_{n-1}))\big)$\;
  }{
    \tcp{Exploitation via value estimation}
    $i_n \leftarrow \arg\max_{i} Q_\theta(s_n, i)$\;
  }

  Evaluate inspiration $i_n$ and obtain reward $r_n$\;

  Update replay memory: $\mathcal{M}_{n+1} \leftarrow \mathcal{M}_{n} \cup \{(i_n, r_n)\}$\;

  Update reflection module with new experience:
  $\text{Reflect}(s_n, r_n) \rightarrow$ contextual summary appended to $s_{n+1}$\;
}
\Return{Final inspiration pool $\{i_n\}$ with associated rewards $\{r_n\}$}
\end{algorithm}


\subsection{Adaptive Reflective Design Exploration}
\label{sec:ARDE}
\myparagraph{Motivation.} 
Typical sampling heuristics in LLMs (\eg, top-$k$, nucleus, or temperature-based decoding) implicitly trade diversity for consistency, but they remain agnostic to empirical feedback from previous designs. 
In addition, designing neural architectures through LLMs entails an inherently \textit{non-differentiable} and \textit{discrete} search space. 
Specifically, unlike gradient-based NAS~\cite{liu2018darts,zhang2021idarts, liu2019dartsdifferentiablearchitecturesearch}, LLM-driven design generation lacks continuous feedback, which often leads to premature convergence toward over-represented and memorized architectural patterns. 
To overcome this limitation, we propose an Adaptive Reflective Design Exploration strategy that adaptively calibrates exploration and exploitation in the non-differentiable, knowledge-driven design space of LLMs.

\myparagraph{Approach.}
Adaptive Reflective Design Exploration explicitly regulates exploration-exploitation through an adaptive $\varepsilon$-greedy schedule~\cite{tokic2010adaptive, rodrigues2009dynamic, dabney2020temporally} that dynamically responds to the uncertainty of recent rewards. 
We reinterpret this idea in a language-conditioned design space, coupling reflective reasoning with an adaptive $\varepsilon$-greedy controller. 
This mechanism allows the agent to self-regulate its curiosity: it increases exploration when joint dynamics is unstable and exploits accumulated knowledge when the search stabilizes.
Let $\mathcal{M}_n = \{(i_{n-1}, r_{n-1})\}$ denote the replay memory containing prior architecture inspiration $i_{n-1}$ and their corresponding scalar rewards $r_{n-1}$. 
At each step $n$, the exploration coefficient is defined, as in the following equation:
\begin{equation}
\varepsilon_n = \varepsilon_{\min} + (\varepsilon_{\max} - \varepsilon_{\min}) 
\cdot \exp\big(-\lambda \cdot \mathrm{Var}(r_{n-m:n})\big),
\label{eq:adaptive_eps}
\end{equation}
where $\mathrm{Var}(r_{n-m:n})$ measures the variance of recent rewards within the latest window $[n-m, n]$. 
Precisely, a lower reward variance encourages \textit{exploration}, while a higher variance signals convergence and induces \textit{exploitation}. 
This dynamic scheduling allows the search policy to adaptively modulate its exploratory intensity without hand-crafted tuning.
Given the current reflective state $s_n$, which summarizes textual memory, prior performance, and contextual embeddings, the agent effectively samples architectural proposals according to the following equation:
\begin{equation}
\small{\pi_n(i \mid s_n) =
\begin{cases}
\arg\max_{i} Q_\theta(s_n, i) \\ \qquad\qquad \text{with probability } 1 - \varepsilon_n, \\[5pt]
\mathcal{A}_{\text{LLM}}\!\left(\mathrm{prompt}(s_n, \mathrm{Reflect}(s_{n-1}, r_{n-1}))\right) \\ \qquad\qquad \text{with probability } \varepsilon_n,
\end{cases}
}\label{eq:reflective_policy}
\end{equation}
where $Q_\theta$ represents a learned value estimator predicting the expected reward for a candidate inspiration $i$ under the current reflective state $s_n$ (details in Appendix~\cref{app:arde}), and $\mathrm{Reflect}(\cdot)$ reformulates the design prompt based on prior reasoning and performance feedback. 
This mechanism allows the system to alternate between exploiting historically strong design cues and exploring new reasoning paths guided by reflective memory.
After evaluation, each structural inspiration $i_n$ receives a scalar reward $r_n$, and the tuple $(i_n, r_n)$ is appended to $\mathcal{M}_{n+1}$. 
$\mathrm{Reflect}$ then summarizes accumulated outcomes and failures to reshape subsequent reasoning trajectories. 
In this way, the agent performs implicit policy improvement without gradient updates; this reflective feedback protocol enhances representational diversity and conceptual depth, while our adaptive module enforces interpretable and variance-regulated exploration.


\begin{algorithm}[t]
\caption{Pareto-guided Evolutionary Selection}
\label{alg:PES}
\KwIn{\\
\qquad Candidate models $\mathcal{A}_t=\{a_t^{(n)}\}$; \\
\qquad Objective vectors $\mathbf{v}(a)=$\\ \qquad\quad[$\text{Acc}(a), \text{Params}(a), \text{Latency}(a),$\\ \qquad\qquad $\text{StructDiv.}(a), \text{Conf.}(a)]$;\\
\qquad Bidding weights $\{\beta,\lambda_p,\lambda_\ell,\gamma_d,\rho_c\}$; \\
\qquad Survival ratio $\kappa\in(0,1]$.
}

\KwOut{\\
\qquad Survivors $\mathcal{S}$ for the next generation.
}

\BlankLine
\textbf{Procedure:}

\BlankLine
\tcp{(1)Pareto front decomposition (fast non-dominated sort)}
$\{\mathcal{F}_0,\mathcal{F}_1,\dots\} \leftarrow \text{FastNonDominatedSort}(\mathcal{A}_t)$

\tcp{(2)Candidate set for bidding: prefer first front}
$\mathcal{C}\leftarrow
\begin{cases}
\mathcal{F}_0, & \text{if }|\mathcal{F}_0|>0\\
\mathcal{A}_t, & \text{otherwise}
\end{cases}$

\tcp{(3)Compute bids on $\mathcal{C}$ (linear penalties)}
\ForEach{$a\in\mathcal{C}$}{
  Estimate $\sigma(a)$ from intermediate training dynamics;\\
  \(\displaystyle
  \text{Bid}(a)=\text{Acc}(a)
  -\lambda_p \text{Params}(a)
  -\lambda_\ell \text{Latency}(a)
  +\gamma_d \text{StructDiv.}(a)
  +\rho_c \text{Conf.}(a)
  +\beta \sigma(a)
  \)
}

\tcp{(4)Pick survivors by bid within $\mathcal{C}$}
$K\leftarrow \max\!\bigl(1,\ \lfloor \kappa\cdot |\mathcal{C}|\rfloor\bigr)$;\\
$\mathcal{S}\leftarrow \operatorname{TopK}\!\bigl(\mathcal{C},K;\ \text{key}=\text{Bid}\bigr)$;

\Return $\mathcal{S}$
\end{algorithm}

\subsection{Pareto-guided Evolutionary Selection}
\label{sec:PES}

\textbf{Motivation.} 
NAD based on iterative propose--and--train cycles often suffers from \emph{mode collapse}, where the design converges prematurely toward architectures that improve a single objective (\eg, accuracy), while independently disregarding efficiency, robustness, or architectural diversity. 
Such collapse often leads to search inefficiency and limits the interpretable design principles in various scenarios (\ie, from lightweight on-device inference to reliable performance under distributional shift).
In this paper, we carefully design a Pareto-Guided Evolutionary Selection strategy that guarantees diverse high-quality candidates while promoting models that achieve balanced trade-offs across performance, efficiency, confidence, and  structural diversity.


\begin{table*}[t]
    \begin{center}
    \resizebox{\linewidth}{!}{%
    \begin{tabular}{m{3.5cm}|C{2cm}|C{2.2cm}C{2.2cm}|C{2.2cm}C{2.2cm}|C{2.2cm}C{2.2cm}}
    \toprule
 \multirow{2.3}{*}{\textbf{Method}}   & \multirow{2.3}{*}{\textbf{$\#$ Archs}} & \multicolumn{2}{c|}{\textbf{CIFAR10}~($\uparrow$)}   & \multicolumn{2}{c|}{\textbf{CIFAR100}~($\uparrow$)}  & \multicolumn{2}{c}{\textbf{ImageNet16-120}~($\uparrow$)} \\
 \cmidrule{3-8}
& & \textbf{Val.} & \textbf{Test} & \textbf{Val.} & \textbf{Test} & \textbf{Val.} & \textbf{Test} \\\midrule
ENAS ~\cite{pham2018efficient}
& - & 37.51±3.19 &  53.89±0.58 & 13.37±2.35 & 13.96±2.33 & 15.06±1.95 & 14.84±2.10 \\
DARTS ~\cite{liu2018darts}                
& - & 39.77±0.00 & 54.30±0.00 & 38.57±0.00 & 15.61±0.00 & 18.87±0.00 & 16.32±0.00 \\
SETN ~\cite{dong2019one}    
& - & 84.04±0.28 & 87.64±0.00 & 58.86±0.06 & 59.05±0.24 & 33.06±0.02 & 32.52±0.21 \\
DSNAS ~\cite{xu2019pc}
& - & 89.66±0.29 & 93.08±0.13 & 30.87±16.40 & 31.01±16.4 & 40.61±0.10 & 41.07±0.09 \\
PC-DARTS ~\cite{xu2019pc}                                                  
& - & 89.96±0.15 & 93.41±0.30 & 67.12±0.39 & 67.48±0.89  & 40.83±0.08 & 41.31±0.22 \\
SNAS ~\cite{xie2018snas}                                               
& - & 90.10±1.04 & 92.77±0.83 & 69.69±2.39 & 69.34±1.98 & 42.84±1.79 & 43.16±2.64 \\
iDARTS~\cite{zhang2021idarts}                                                    
& - & 89.86±0.60 & 93.58±0.32 & 70.57±0.24 & 70.83±0.48 & 40.38±0.59 & 40.89±0.68 \\
GDAS  ~\cite{dong2019searching}                                                
& - & 89.89±0.08 & 93.61±0.09& 71.34±0.04 & 70.70±0.30 & 41.59±1.33 & 41.71±0.98 \\
DRNAS ~\cite{chen2020drnas}                                                  
& - & 91.55±0.00 & 94.36±0.00 & 73.49±0.00 & 73.51±0.00 & 46.37±0.00 & 46.34±0.00 \\
$\beta$-DARTS ~\cite{ye2022b}                                                  
& - & 91.55±0.00 & 94.36±0.00 & 73.49±0.00 & 73.51±0.00 & 46.37±0.00 & 46.34±0.00 \\
$\Lambda$-DARTS  ~\cite{movahedi2022lambda}                                               
& - & 91.55±0.00 & 94.36±0.00 & 73.49±0.00 & 73.51±0.00 & 46.37±0.00 & 46.34±0.00 \\
REA ~\cite{dong2020bench}                                                  
& 500 & 91.19±0.31 & 93.92±0.30 & 71.81±1.12 & 71.84±0.99 & 45.15±0.89 & 45.54±1.03 \\
RS  ~\cite{real2019regularized}                                          
& 500 & 90.93±0.36 & 93.70±0.36 & 70.93±1.09 & 71.04±1.07 & 44.45±1.10 & 44.57±1.25 \\
REINFORCE ~\cite{williams1992simple}
& 500 & 91.09±0.37 & 93.85±0.37 & 71.61±1.12 & 71.71±1.09 & 45.05±1.02 & 45.24±1.18 \\
BOHB ~\cite{falkner2018bohb}                                                  
& 500 & 90.82±0.53 & 93.61±0.52 & 70.74±1.29 & 70.85±1.28 & 44.26±1.36 & 44.42±1.49 \\
\textit{Oracle} & - & 91.61 & 94.37 & 73.49 & 73.51 & 46.77 & 47.31 \\
\cmidrule{1-8}
GENIUS ~\cite{zheng2023can}                                               
& 10 & 91.07±0.20 & 93.79±0.09 & 70.96±0.33 & 70.91±0.72 & 45.29±0.81 & 44.96±1.02 \\
LLMatic ~\cite{nasir2024llmatic}                                                
& 2000 & - & 94.26±0.13 & - & 71.62±1.73 & - & 45.87±0.96 \\
\cmidrule{1-8}
LeMo-NADe-GPT4 ~\cite{rahman2024lemo}
& 30 & 90.90 & 89.41 & 63.38 & 67.90 & 27.05 & 27.70 \\
NADER ~\cite{yang2025nader}
& 5 & 91.17±0.24 & 94.52±0.22 & 73.29±1.86 & 73.12±1.09 & 47.98±0.73 & 47.99±0.38 \\
NADER ~\cite{yang2025nader}
& 10 & 91.18±0.23 & 94.52±0.22 & 74.71±0.45 & 74.65±0.33 & 48.56±0.83 & 48.61±0.76 \\
NADER ~\cite{yang2025nader}
& 500 & 91.55 & 94.62 & 75.72 & 76.00 & 50.20 & 50.52 \\
\cmidrule{1-8}
\coloredrowcell{DCDCDC}\coloredcell{DCDCDC}
RevoNAD (Qwen2.5)
& 5 & 91.17±0.25 & 94.77±0.18 & 76.21±0.42 & 76.25±0.37 & \textbf{51.00±0.21} & 50.55±0.17 \\
\coloredrowcell{DCDCDC}\coloredcell{DCDCDC}
RevoNAD (GPT4o)
& 5 & \textbf{92.55±0.27} & \textbf{95.22±0.20} & \textbf{76.70±0.39} & \textbf{76.38±0.32} & 50.58±0.19 & \textbf{50.72±0.20} \\
\bottomrule
\end{tabular}}
\vspace{-1em}
\caption{Comparison with existing state-of-the-art NAS and NAD methods on CIFAR10, CIFAR100 and ImageNet16-120. Averaged over 4 independent search runs (\ie, mean±std). The \textbf{bold} values indicate the best performance. \colorbox{gray!20}{Gray shading} indicates Ours.}
\label{tab:main}
\end{center}
\end{table*}


\myparagraph{Approach.}
At generation $t$, a proposal pool $\mathcal{A}_t$ is produced by a set of collaborating generative design and reflective exploration agents.
In particular, each proposed model $a_{t}^{(n)}$ is generated from parents $a_{t-1}^{(n)}$ and an inspiration token $i_t^{{(n)}}$:
\begin{equation}
    a_{t}^{(n)} = \mathcal{T}(a_{t-1}^{(n)}, i_t^{(n)}), 
    \qquad i_t^{(n)} \sim \pi_t(i \mid s_t^{(n)}),
\end{equation}
where $\mathcal{T}$ denotes a compositional architectural transformation and $\pi_t(\cdot)$ is the evolving inspiration distribution.
Additionally, each candidate model $a_{t}^{(n)}$ is evaluated along a vector of objectives:
$\mathbf{v}(a) = [\text{Acc}(a), \text{Params}(a), \text{Latency}(a)$, $\text{StructDiv.}(a), \text{Conf.}(a)] \in \mathbb{R}^5$,
where accuracy-related objectives are maximized, while parameter count and latency are minimized to ensure both performance and deployability.
First, we apply fast non-dominated sorting to decompose the model population into Pareto fronts
$\{\mathcal{F}_0, \mathcal{F}_1, \dots\} = \text{FastNonDominatedSort}(\mathcal{A}_t)$,
where $\mathcal{F}_0$ denotes the set of models not dominated by any other candidate.
In practice, if $\mathcal{F}_0$ is non-empty, we restrict selection to $\mathcal{F}_0$; otherwise we fall back to all candidates, as in the following equation:
\begin{equation}
\mathcal{C} =
\small{\begin{cases}
\mathcal{F}_0, & \text{if } |\mathcal{F}_0| > 0,\\
\mathcal{A}_t, & \text{otherwise.}
\end{cases}
}\end{equation}
Beyond Pareto feasibility and geometric spread in objective space, we introduce a bidding score that integrates practical constraints and transferability considerations:
\begin{equation}
\label{eq:bid}
\small
\begin{aligned}
\text{Bid}(a) ={}& \text{Acc}(a)
- \lambda_p\!\cdot\!\text{Params}(a)
- \lambda_\ell\!\cdot\!\text{Latency}(a)\\[-0.15em]
&+ \gamma_d\!\cdot\!\text{StructDiv.}(a)
+ \rho_c\!\cdot\!\text{Conf.}(a)
+ \beta\!\cdot\!\sigma(a),
\end{aligned}
\end{equation}
where $\sigma(a)$ denotes performance uncertainty estimated from intermediate training dynamics, and $\{\lambda_p, \lambda_\ell, \gamma_d, \rho_c, \beta\}$ control trade-offs between efficiency,  structural diversity, confidence, and exploration (see Appendix~\cref{app:pes}).
Models are prioritized first by Pareto front index, and then by bidding score:
$(\text{Front Index}\downarrow,\;\text{Bid}(a)\uparrow)$,
favoring Pareto-optimal candidates with preserved diversity and strong practicality. 
Let $\kappa \in (0,1]$ be the survival ratio.
We rank $\mathcal{C}$ by $\text{Bid}(a)$ in descending order and select the top $\lfloor \kappa|\mathcal{C}| \rfloor$ survivors:
\begin{equation}
\mathcal{S}_t = \texttt{Top}_{\lfloor \kappa|\mathcal{C}| \rfloor}\;\text{by}\;\text{Bid}(a).
\end{equation}
This procedure stabilizes the evolutionary process, prevents degeneration toward overly specialized architectures, and maintains a consistent supply of structurally novel yet performant candidates.
Subsequently, we apply Pareto-first bidding (Algorithm~\ref{alg:PES}) to select survivors for the next generation, preserving diversity while guiding performance-aware progression toward more generalizable architectural motifs.


%% file: sec/4_exp.tex
\section{Experiments}

\subsection{Experimental setup.}

\myparagraph{Dataset.}
We follow the official NAS-Bench-201~\cite{dong2020bench} training and evaluation protocols, 
covering CIFAR10, CIFAR100~\cite{krizhevsky2009learning}, ImageNet16-120~\cite{chrabaszcz2017downsampled}, COCO-5K~\cite{lin2014microsoft}, and Cityscapes~\cite{Cordts2016Cityscapes}.
RevoNAD adopts a ResNet-18~\cite{he2016deep} architecture as the base structure and iteratively refines it through the adaptive reflective generation strategy.
Additionally, we adopt YOLACT~\cite{bolya2019yolact} and SemanticFPN~\cite{kirillov2019panoptic} as base models to evaluate the scalability of our approach to Object Detection and Semantic Segmentation, respectively.
All experiments are performed over 3 generations, each conducting 5 architecture searches. 
Further implementation details and dataset configurations are provided in Appendix~\cref{app:impl}.

\myparagraph{Baseline.}
We first compare RevoNAD with representative NAS approaches, including random search-based~\cite{bergstra2012random,li2020random}, evolutionary, RL-based~\cite{pham2018efficient,williams1992simple}, HPO-based~\cite{falkner2018bohb}, and differentiable algorithms~\cite{liu2018darts,dong2019one,hu2020dsnas,xu2019pc,xie2018snas,zhang2021idarts,dong2019searching,chen2020drnas,ye2022b,movahedi2022lambda,xu2019pc}, as well as recent LLM-assisted NAS frameworks~\cite{zheng2023can,nasir2024llmatic}. 
We further compare against LLM-based NAD frameworks such as LeMo-NADe~\cite{rahman2024lemo} and NADER~\cite{yang2025nader}, which propose architectural designs leveraging LLM-based agents.

\subsection{Overall Performance}
\myparagraph{Comparison with existing SOTA approaches.}
In this section, we quantitatively compares RevoNAD to representative NAS and NAD approaches on CIFAR10/100, and ImageNet16-120 leveraging NAS-Bench-201~\cite{dong2020bench} in~\cref{tab:main}. 
While traditional NAS methods (\ie, BOHB, REINFORCE, RS, REA) require large-scale search over hundreds of candidate architectures, and recent LLM-guided methods (\ie, GENIUS, LeMo-NADe, NADER) still rely on extensive sampling or specialized retraining pipelines, RevoNAD (GPT4o) achieves superior results with significantly fewer designed architectures. 
With only 5 generated architectures (\ie, \# Archs), RevoNAD further improves to 92.55/95.22 on CIFAR10, 76.70/76.38 on CIFAR100, and 50.58/50.72 on ImageNet16-120, establishing a new state-of-the-art for NAD under comparable or lower search cost. 
These results highlight that RevoNAD does not rely on large search budgets; instead, its reflective multi-expert consensus and Pareto-guided evolutionary refinement enable rapid discovery of high-quality architectures with minimal exploration overhead compared to prior NAD/NAS methods.

\begin{figure*}[t]
\centering
\includegraphics[width=1\linewidth]{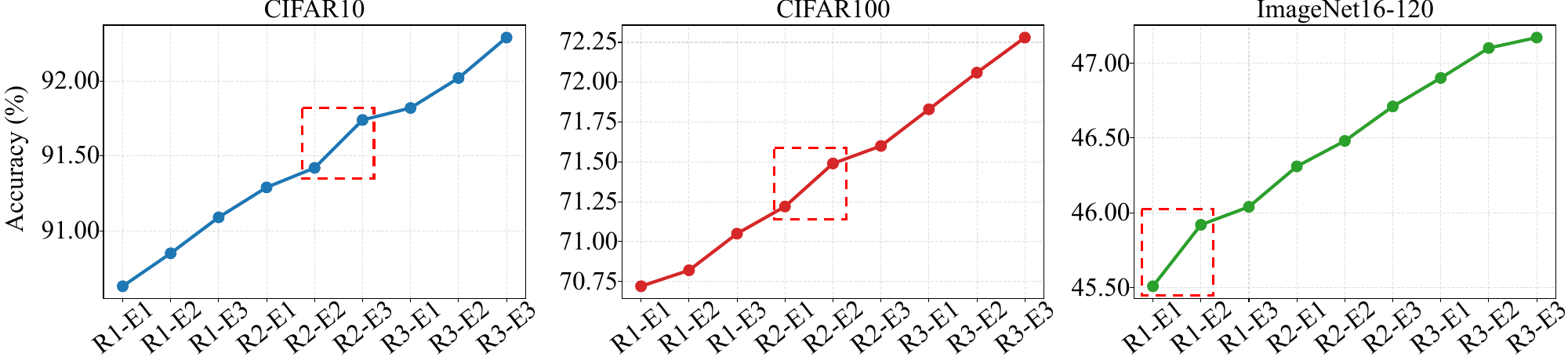}
\vspace{-2em}
\caption{Multi-round Multi-expert Consensus refinement across varying numbers of experts and refinement rounds. R1–E2 denotes 1 round and 2 experts, respectively. Note that the red dotted box highlights the most significant improvement.}
\label{fig:abl_mmc}
\vspace{-1em}
\end{figure*}

\subsection{Ablation Study}
\myparagraph{Effect of Individual Modules.}
We quantitatively analyze the contribution of the two key components in RevoNAD: Adaptive Reflective Design Exploration (ARDE) and Pareto-guided Evolutionary Selection (PES). As shown in 
~\cref{tab:ablation core}, removing both modules leads to substantially weaker performance across all datasets, indicating that naive single-shot LLM proposals are insufficient for stable architecture discovery. Introducing PES alone improves results by filtering inferior candidates through multi-objective selection, while enabling ARDE alone yields consistent gains by refining architectural motifs along concept-aligned expert axes. Combining both ARDE and PES achieves the highest performance, improving CIFAR10 test accuracy from 92.31\% to 95.22\%, CIFAR100 from 71.83\% to 76.38\%, and ImageNet16-120 from 46.90\% to 50.72\%. These results demonstrate that ARDE and PES are complementary and jointly essential for stable and effective LLM-guided neural architecture design.

\myparagraph{Impact of Multi-Round Iterations and Expert Pool Size.}
In this work, we observe that single-shot LLM-based architectural ideation often lacks sufficient refinement and verification, which can lead to hallucinated or semantically inconsistent design cues. 
To ensure a fair comparison, we fix \texttt{resnet\_basic} as the parent architecture and perform one design generation per setting, measuring performance under varying numbers of discussion rounds and expert pool sizes. As shown in~\cref{fig:abl_mmc}, both increasing discussion rounds and enlarging the sub-axis expert pool improve performance; however, it is noteworthy the dominant gain (red dotted boxes) comes from adding more sub-axis experts, indicating that \textit{diverse conceptual perspectives provide more effective validation and refinement}. This empirically supports that multi-perspective collaborative discussion is essential for producing stable, informative architectural clues.

\begin{table}[t]
    \begin{center}
    \resizebox{\linewidth}{!}{%
    \begin{tabular}{cc|cc|cc|cc}
    \toprule
 \multicolumn{2}{c|}{\textbf{Method}} & \multicolumn{2}{c|}{\textbf{CIFAR10}~($\uparrow$)}   & \multicolumn{2}{c|}{\textbf{CIFAR100}~($\uparrow$)}  & \multicolumn{2}{c}{\textbf{ImageNet16-120}~($\uparrow$)} \\
 \midrule
 \textbf{ARDE} & \textbf{PES} & \textbf{Val.} & \textbf{Test} & \textbf{Val.} & \textbf{Test} & \textbf{Val.} & \textbf{Test} \\
 \midrule
 \ding{55} & \ding{55} &
 88.43±0.37 & 92.31±0.51 & 71.95±0.34 & 71.83±0.26 & 46.67±0.50 & 46.90±0.29 \\
 \midrule
  \ding{51}& \ding{55} & 
  92.09±1.28 & 94.66±1.05 & 75.70±0.44 & 75.74±0.37 & 49.60±0.37 & 49.78±0.87 \\
  \ding{55}&\ding{51}&
  90.61±0.45 & 93.97±0.60 & 73.16±0.14 & 73.49±0.18 & 47.23±0.36 & 47.62±0.34 \\
 \midrule
\ding{51}&\ding{51}&
 \textbf{92.55±0.27} & \textbf{95.22±0.20} & \textbf{76.70±0.39} & \textbf{76.38±0.32} & \textbf{50.58±0.19} & \textbf{50.72±0.20} \\
\bottomrule
\end{tabular}}
\vspace{-1em}
\caption{Ablation study of the core components in RevoNAD.}
\label{tab:ablation core}
\end{center}
\vspace{-2em}
\end{table}

\begin{figure*}[t]
\centering
\includegraphics[width=1\linewidth]{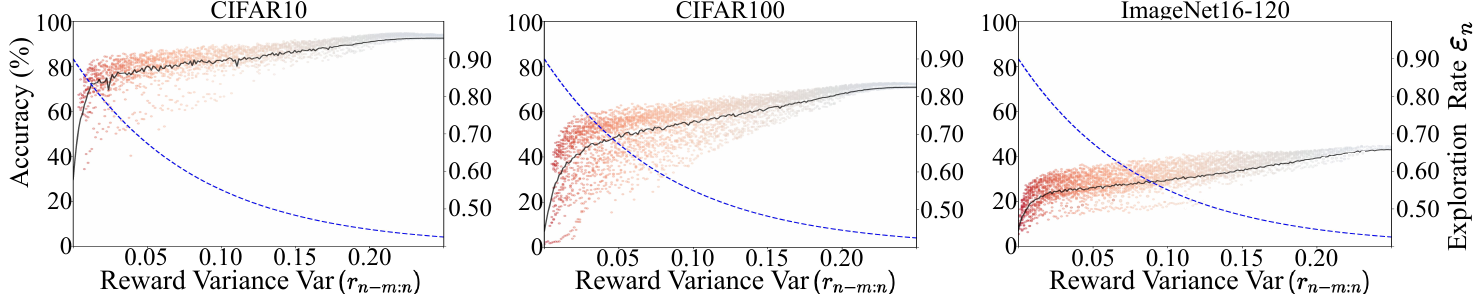}
\vspace{-1em}
\caption{Adaptive exploration schedule. High joint dynamics of $\varepsilon_n$ encourages exploration, while low variance drives exploitation.}
\label{fig:figure_eps_ratio}
\end{figure*}

\subsection{Design Dynamics}

\myparagraph{Interpretability via Parameter Variation.}
We quantitatively investigate the impact of Pareto preference weights on architecture selection to understand how different optimization priorities influence both architectural structure and deployment characteristics. 
Specifically, we vary the weighting coefficients for parameter cost ($\lambda_p$), confidence regularization ($\rho_c$), and latency constraint ($\lambda_l$), while keeping all other training conditions identical, as shown in~\cref{tab:pareto}.
Increasing $\lambda_p$ encourages more compact models with reduced computational cost, achieving 1.33M parameters and 0.188 GFLOPs while maintaining comparable accuracy. 
A higher confidence weight $\rho_c$ yields the most accurate configuration (75.3\%), suggesting that reliability-aware weighting produces more generalizable and consistent architectures, as detailed in the Appendix~\cref{app:exp}. 
In contrast, emphasizing latency ($\lambda_l=0.25$) effectively reduces inference time (16.8ms) but slightly compromises accuracy, reflecting the inherent trade-off between computational efficiency and predictive precision.
These results indicate that our Pareto-based multi-objective formulation provides a flexible mechanism to navigate trade-offs among accuracy, efficiency, and model compactness. 
Ultimately, the ability to tune these preference weights not only enhances the interpretability of architectural design decisions—clarifying how each factor contributes to performance—but also facilitates adaptive deployment across diverse computing environments, from lightweight edge devices to high-throughput GPU servers.

\begin{table}[t]
    \begin{center}
    \resizebox{\linewidth}{!}{%
    \begin{tabular}{ccc|cccc}
    \toprule
 \multicolumn{3}{c|}{\textbf{Pareto}} & \multicolumn{4}{c}{\textbf{CIFAR100}}  \\
 \midrule
 \textbf{Params $\lambda_p$} & \textbf{Conf. $\rho_c$} & \textbf{Latency $\lambda_{l}$} & \textbf{\# Params} & \textbf{GFLOPs} & \textbf{Latency} & \textbf{Acc.}\\
 \midrule
 0.10 & 0.10 & 0.10 
 & 1.39M & 0.193 & 23.6ms & \textbf{75.80}  
 \\
 0.25 & 0.10 & 0.10 
 & \textbf{1.33M} & \textbf{0.188} & 22.7ms & 74.32 \\
 0.10 & 0.25 & 0.10 
 & 1.41M & 0.193 & 17.6ms & 75.30 \\
 0.10 & 0.10 & 0.25
 & 1.34M & \textbf{0.188} & \textbf{16.8ms} & 73.21 \\
\bottomrule
\end{tabular}}
\vspace{-1em}
\caption{Effect of Pareto preference weights on architecture selection. Further experiments are provided in Appendix~\cref{app:exp}.}
\label{tab:pareto}
\end{center}
\vspace{-1em}
\end{table}

\myparagraph{Exploration–Variance–Performance Relationship.}
In this study, we showcase how reward variance $\mathrm{Var}(r_{n-m:n})$ and exploration rate $\varepsilon_n$ 
affect the performance of architectural design, as shown in~\cref{fig:figure_eps_ratio}. 
In particular, we analyze the joint dynamics of up to 500 samples for each dataset (\ie, CIFAR10, CIFAR100, and ImageNet16-120).
For better understanding, we plot the reference trajectory of $\varepsilon_n$, showing the averaged trend across multiple runs and its convergence behavior.
Note that outliers are removed using Z-score filtering ($z < 2.5$), and adaptive Gaussian noise ($\sigma_t = \sigma_{\max} e^{-kt}$) is applied to visualize the variance envelope around the accuracy curve.  
Our results indicate that lower reward variance (\ie, exploration coefficient $\varepsilon_n\uparrow$) induce significant accuracy oscillations in the early stages; however, as training proceeds, joint dynamics of $\varepsilon_n$ diminish, yielding smoother and more stable evolutionary convergence.
Overall, these findings demonstrate that maintaining an appropriate balance between exploration and reward variance leads to more stable convergence, while facilitating a more interpretable and efficiently expandable architectural design space.

\subsection{Architectural Behavior}
\myparagraph{Success Rate of Pareto-guided Model Generation.}
While expanding the search space in NAS and NAD can increase architectural diversity, it also introduces substantial computational overhead, making effective space reduction essential. In LLM-guided NAD settings, the key lies in generating constructive child architectures from a given pareto-guided parent, rather than sampling blindly from a large search space. However, prior approaches such as NADER evaluate models primarily by accuracy, offering limited insight into model capacity. In contrast, RevoNAD employs a Pareto-based evaluation to jointly assess accuracy, efficiency, and robustness, enabling more reliable evolutionary selection and stable search-space pruning. In~\cref{tab:succ_rate}, our Pareto-guided strategy improves the parent-to-child positive generation rate by +24.9\%, achieves valid positive offspring in as few as 1.4 trials, and requires on average only 2.3 attempts to obtain a positive pair. These results demonstrate that Pareto-based selection provides a principled criterion for architectural viability, leading to more stable and scalable NAD.

\begin{table}[t]
\centering
\small
\resizebox{\linewidth}{!}{%
\begin{tabular}{p{1.5cm}|ccc}
\toprule
\textbf{Method} & \textbf{Succ. Rate} ($\uparrow$) & \textbf{Trials-to-1st-Succ.} ($\downarrow$) & \textbf{Trials-per-Succ.} ($\downarrow$) \\
\midrule
NADER        & 22.6\% & 3.7 & 4.8 \\
\rowcolor[HTML]{DCDCDC} 
RevoNAD      & \textbf{47.5\%} & \textbf{1.4} & \textbf{2.3} \\
\bottomrule
\end{tabular}}
\vspace{-1em}
\caption{
Success rate and trial efficiency in parent–child model generation.
This table reports how frequently child models, generated from parent architectures, 
lead to improved performance (\textit{Success Rate}), 
along with the average number of attempts required to obtain the first successful model 
(\textit{Trials-to-1st-Success}) and the mean number of trials per successful generation 
(\textit{Trials-per-Success}). }
\label{tab:succ_rate}
\end{table}

\myparagraph{Transferability and Generalization.}
To assess the transferability of our framework, we evaluate RevoNAD under heterogeneous downstream configurations, including object detection and instance segmentation (COCO-5K) and semantic segmentation (Cityscapes). 
In each setting, we integrate our orchestrator on existing methods (YOLACT and SemanticFPN) while keeping the original training protocols identical. 
These results are significant in demonstrating the \emph{general applicability and latent potential} of our approach beyond classification domain, as shown in~\cref{tab:transfer}.
Specifically, RevoNAD achieves consistent improvements across different tasks and metrics, indicating that the same sub-axis expert configuration transfers effectively from image classification to object detection and semantic segmentation without re-alignment. This suggests that the decomposition captures task-general architectural principles rather than surface-level heuristics.
Furthermore, Adaptive Reflective Design Exploration (ARDE) and Pareto-guided Evolutionary Selection (PES), empirically demonstrate their ability to provide reliable search directions and maintain stable performance under varying optimization regimes.
Ultimately, this qualitative implication is substantial: our orchestrator shows strong interpretability in its design behaviors and adaptability across tasks, confirming its potential as a unified and extensible architectural reasoning framework for various applications.

\begin{table}[t]
\centering
\label{tab:transfer}
\begin{subtable}{\linewidth}
\centering
\resizebox{\linewidth}{!}{
\begin{tabular}{p{3cm}|C{2.3cm}C{2.3cm}C{2.3cm}}
\toprule
\multirow{2}{*}{\textbf{Method}} & \multicolumn{3}{c}{\textbf{COCO-5K}} \\
\cmidrule(lr){2-4}
& $\mathbf{AP}_{\text{bbox}}$~($\uparrow$) 
& $\mathbf{AP@50}_{\text{bbox}}$~($\uparrow$)
& $\mathbf{AP}_{\text{mask}}$~($\uparrow$) \\
\midrule
YOLACT  & 13.5 & 28.9 & 12.6 \\
\rowcolor[HTML]{DCDCDC} 
w/ RevoNAD & \textbf{13.7} & \textbf{29.6} & \textbf{12.8} \\
\bottomrule
\end{tabular}}
\small\caption{Object Detection and Instance Segmentation.}
\end{subtable}

\vspace{0.4em}

\begin{subtable}{\linewidth}
\centering
\resizebox{\linewidth}{!}{
\begin{tabular}{p{3cm}|C{2.3cm}C{2.3cm}C{2.3cm}}
\toprule
\multirow{2}{*}{\textbf{Method}} & \multicolumn{3}{c}{\textbf{Cityscapes}} \\
\cmidrule(lr){2-4}
& $\mathbf{aAcc}$~($\uparrow$) & $\mathbf{mAcc}$~($\uparrow$)& $\mathbf{mIoU}$~($\uparrow$) \\
\midrule
SemanticFPN  & 95.83 & 81.80 & 74.01 \\
\rowcolor[HTML]{DCDCDC} 
w/ RevoNAD & \textbf{95.88} & \textbf{82.73} & \textbf{75.05} \\
\bottomrule
\end{tabular}}
\small{\caption{Semantic Segmentation.}}
\end{subtable}
\vspace{-2em}
\caption{Transfer studies on (a) COCO-5K and (b) Cityscapes.}
\end{table}

%% file: sec/5_con.tex
\section{Conclusion}
\label{sec:con}

\myparagraph{Summary.}
In this work, we proposed RevoNAD, a reflective evolutionary framework for Neural Architecture Design (NAD) that shifts focus from low-level hyperparameter search to global architectural reasoning about high-level structural principles. RevoNAD integrates multi-round multi-expert consensus to accumulate coherent architectural knowledge, adaptive reflective exploration to align exploration and refinement based on reward variance, and Pareto-guided evolutionary selection to ensure both performance and structural diversity. Across CIFAR10/100, ImageNet16-120, COCO-5K, and Cityscapes, RevoNAD consistently achieves superior results while producing architectures that are interpretable and practically deployable. Finally, this finding suggests that neural architecture design benefits significantly from a feedback-aligned iterative reasoning, providing a scalable path toward robust and optimal model search.

\myparagraph{Limitations.}
RevoNAD’s architectural insights ultimately depend on the underlying LLM’s knowledge capacity.
The evolutionary refinement loop still requires meaningful training feedback, which can incur computational cost at scale.
We view addressing these challenges (\ie, particularly stronger domain adaptation and compute-efficient training feedback) as valuable directions for future work.

\section*{Acknowledgement}
This work was supported by
Culture, Sports and Tourism R\&D Program through the Korea Creative Content Agency grant funded by the Ministry of Culture, Sports and Tourism 
(International Collaborative Research and Global Talent Development for the Development of Copyright Management and Protection Technologies for Generative AI, RS-2024-00345025, 25\%;
Research on neural watermark technology for copyright protection of generative AI 3D content, RS-2024-00348469, 35\%), 
the National Research Foundation of Korea(NRF) grant funded by the Korea government(MSIT)(RS-2025-00521602, 28\%),
Institute of Information \& communications Technology Planning \& Evaluation (IITP) \& ITRC(Information Technology Research Center) grant funded by the Korea government(MSIT) (No.RS-2019-II190079, Artificial Intelligence Graduate School Program(Korea University), 1\%; 
IITP-2025-RS-2024-00436857, 1\%),
Electronics and Telecommunications Research Institute(ETRI) grant funded by the Korean government(25ZC1200, Research on hyper-realistic interaction technology for five senses and emotional experience, 10\%),
and
Artificial intelligence industrial convergence cluster development project funded by the Ministry of Science and ICT(MSIT, Korea)\&Gwangju Metropolitan City.

%% file: sec/X_suppl.tex
\clearpage
\setcounter{page}{1}
\maketitlesupplementary

\appendix


\appendixsetup
\appendixtableofcontents

\section{Detailed Formulation and Implementation}
\label{app:impl}

\subsection{Implementation details}
\label{app:implementation details}
All experiments follow the official NAS-Bench-201 training and evaluation settings. 
Each model is trained for 200 epochs using stochastic gradient descent (momentum 0.9, weight decay $5\times10^{-4}$) with a cosine-annealed learning-rate schedule starting at 0.1. 
RevoNAD adopts a ResNet architecture as the base structure and is additionally evaluated on randomly sampled architectures from NAS-Bench-201 to verify robustness. The design process consists of 3 generations, each producing 5 architecture candidates guided by reinforcement feedback to iteratively refine network designs. 
All language-model–driven experiments are executed on a workstation equipped with four NVIDIA RTX A6000 GPUs.

\subsection{Language model configuration}
All multi-agent components (\textit{Planner}, \textit{Designer}, \textit{Evaluator}) employ GPT-4o as the primary underlying large language model. 
Each agent communicates via structured prompts that encode the current design objectives, architectural constraints, and performance feedback from previously trained models. 
Unless otherwise specified, we standardize the decoding and context settings across all experiments, using a maximum context length of 4096 tokens, a temperature of 0.7, nucleus sampling with $p=0.9$, and truncating each response to 512 tokens. 
In addition to GPT-4o, we also utilize open-source large language models, including Qwen2.5 and DeepSeek, for selected experiments (\ie, cost- or latency-sensitive settings); unless otherwise noted, these models share the same decoding configuration as described above.

\subsection{Notation Recap}

\begin{table}[h!]
\centering
\begin{tabular}{lp{0.65\linewidth}}
\toprule
\textbf{Symbol} & \textbf{Description} \\
\midrule
$\mathcal{A} = \{a^{(n)}\}$ & Architecture search space; set of all candidate architectures. \\
$a_t^{(n)}$ & $n$-th architecture at generation $t$. \\
$\mathcal{I}_t = \{i_t^{(n)}\}$ & Inspiration set used at generation $t$. \\
$\mathcal{X} = \{x_1,\dots,x_K\}$ & Set of architecture-influential conceptual axes extracted from $P$. \\
$E_k$ & $k$-th expert LLM agent specialized in axis $x_k$. \\
$\Omega$ & Consensus module that merges expert proposals into $\mathcal{I}_t$. \\
\midrule
$\mathcal{M}_n$ & Replay / reflection memory at step $n$. \\
$s_n$ & Reflective state at step $n$ (text, rewards, context embeddings). \\
$\varepsilon_n$ & Exploration coefficient at step $n$ in the adaptive $\varepsilon$-greedy policy. \\
$r_{n-m:n}$ & Recent reward window from step $n-m$ to $n$. \\
$\pi_n(i \mid s_n)$ & Inspiration-sampling policy at step $n$ given state $s_n$. \\
$\mathcal{A}_{\text{LLM}}$ & LLM-based generator that outputs new inspirations from prompts. \\
$Q_\theta(s,i)$ & Value estimator (Q-network) with parameters $\theta$; predicts expected reward of inspiration $i$ in state $s$. \\
\midrule
$\sigma(a)$ & Performance uncertainty of $a$ estimated from intermediate training dynamics. \\
$\mathcal{F}_0$ & First (non-dominated) Pareto front. \\
$\{\mathcal{F}_0,\mathcal{F}_1,\dots\}$ & Pareto fronts obtained by fast non-dominated sorting of $\mathcal{A}_t$. \\
$\mathrm{Bid}(a)$ & Scalar bidding score combining accuracy, efficiency, structual diversity, confidence, and uncertainty. \\
$\mathcal{S}_t$ & Survivor set at generation $t$ selected for the next evolutionary cycle. \\
\bottomrule
\end{tabular}
\caption{Notation recap for the RevoNAD framework.}
\label{tab:notation}
\end{table}

\subsection{Datasets.}
\label{app:datasets}
We evaluate RevoNAD on five datasets. 
The first three follow the NAS-Bench-201 benchmark setting for image classification, 
while \textbf{COCO-5K} and \textbf{Cityscapes} are additionally included to examine the scalability and transferability of RevoNAD to object detection and semantic segmentation tasks, respectively.
For classification datasets, we apply standard preprocessing with per-channel normalization, random cropping, and horizontal flipping, 
whereas COCO-5K is processed with per-channel normalization, aspect-ratio–preserving resizing, and random horizontal flipping, following common detection practices. For Cityscapes, we follow standard semantic segmentation preprocessing with per-channel normalization, multi-scale resizing, and random horizontal flipping.
\begin{itemize}
    \item \textbf{CIFAR10:} 60K $32\times32$ color images from 10 classes. 
    The 50K training images are evenly divided into 25K for training and 25K for validation, 
    while the 10K test images remain unchanged.
    
    \item \textbf{CIFAR100:} Same resolution as CIFAR10 but with 100 fine-grained categories. 
    The 10K test set is evenly split into 5K for validation and 5K for testing.
    
    \item \textbf{ImageNet16-120:} A downsampled version of ImageNet16$\times$16, 
    containing 151.7K training, 3K validation, and 3K test images across 120 classes.
    
    \item \textbf{COCO-5K:} A subset of the COCO dataset, comprising 5K training and 5K validation images with bounding-box and instance-mask annotations across 80 object classes.

    \item \textbf{Cityscapes:} A high-resolution urban scene dataset with 19 evaluation classes. It contains 2,975 training, 500 validation, and 1,525 test images providing fine-grained pixel-level semantic annotations.
\end{itemize}

\section{More details of RevoNAD}
\label{app:RevoNAD}

\subsection{Multi-Round Multi-expert Consensus}
\label{app:mmc}

\begin{table*}[t]
\centering
\small
\renewcommand{\arraystretch}{1.15}
\begin{tabular}{p{0.2\linewidth} p{0.8\linewidth}}
\toprule
\textbf{Prompt Type} & \textbf{Prompt Description} \\
\midrule
\textbf{PROMPT\_SUBTASKS} &
\textbf{Role:} Senior computer-vision architect. \newline
\textbf{Goal:} Analyze the paper and extract \emph{core architectural signals} that guide expert agents in decomposing it into design-related sub-tasks. \newline
\textbf{STRICT JSON output:}
\begin{itemize}
  \item \texttt{"tasks"}: [main CV tasks (e.g., detection, segmentation, 3D reconstruction, OCR, SLAM, BEV)]
  \item \texttt{"sub\_tasks"}: [$\le$4 architecture-influential axes (e.g., hierarchy design, receptive-field control, routing/sparsity, normalization layout, parameter sharing)]
  \item \texttt{"keywords"}: [$\le$5 ops or motifs (e.g., deformable conv, attention window, SE block, MoE router, adapter, residual gating)]
\end{itemize}
\textbf{Guidelines:} Focus on architectural levers (hierarchy, connectivity, mixing, routing, memory, inductive bias, interface, stability); avoid dataset or training details; keep concise and transferable. \\
\midrule
\textbf{PROMPT\_EXPERT} &
\textbf{Role:} Expert agent for a specific sub-task (\texttt{"field"}). \newline
\textbf{Goal:} Propose one concise, local, drop-in architectural improvement related to the given field, applicable across CV pipeline stages (e.g., encoding, multi-scale aggregation, temporal/geometric reasoning, cross-modal integration, output mapping). \newline
\textbf{Context:}
\begin{itemize}
  \item \texttt{PAPER (numbered):} {paper\_numbered}
  \item \texttt{CURRENT INSPIRATIONS:} {current\_insp}
\end{itemize}
\textbf{STRICT JSON output:}
\begin{itemize}
  \item \texttt{"proposal"}: $\le$40-word architectural recipe (e.g., DWConv$\rightarrow$SE$\rightarrow$MLP, deformable attention with shift)
  \item \texttt{"rationale"}: $\le$40 words explaining performance/efficiency/generalization benefit
  \item \texttt{"evidence\_refs"}: [supporting sentence indices]
\end{itemize}
\textbf{Guidelines:} Be component-agnostic; focus on architecture-level reasoning; refine or merge if similar ideas exist, otherwise propose a novel one. \\
\midrule
\textbf{PROMPT\_MERGE} &
\textbf{Role:} Orchestrator agent integrating expert proposals. \newline
\textbf{Goal:} Merge, deduplicate, and refine expert JSON proposals into a concise, diverse shortlist of architectural inspirations. \newline
\textbf{Input:} Expert JSON proposals (array of JSON objects). \newline
\textbf{STRICT JSON output:}
\begin{itemize}
  \item \texttt{"inspirations"}: [final concise architectural ideas ($\le$40 words each, up to 4 items)]
\end{itemize}
\textbf{Guidelines:} Merge compatible ideas succinctly; remove redundancy; ensure diversity across scale, efficiency, temporal, and modality reasoning; keep each item modular and transferable. \\
\bottomrule
\end{tabular}
\caption{\textbf{Prompt design for multi-agent architectural ideation.}
Each prompt defines a distinct agent role within our expert-consensus pipeline, balancing diversity and conciseness in architectural reasoning.}
\label{tab:prompts}
\end{table*}

\paragraph{Prompt Design for Multi-Agent Inspiration Extraction}
The multi-round consensus framework relies critically on \emph{role-specialized prompting}
to ensure that textual architectural reasoning is decomposed, contextualized, and refined
in a structured way. Rather than prompting a single LLM instance to extract architectural ideas
in one pass, we define three complementary prompt types corresponding to the three core roles
in MMC: (1) sub-task decomposition, (2) expert proposal, and (3) proposal integration.
Here, \cref{tab:prompts} summarizes these prompts. We highlight several key design principles:

\begin{itemize}
    \item \textbf{Decomposition before Ideation:} PROMPT\_SUBTASKS forces the LLM to first identify
    \emph{architecture-influential axes} (\eg, hierarchy depth, spatial aggregation strategy,
    skip-connectivity, feature routing), preventing single-perspective inspiration collapse.
    
    \item \textbf{Locality and Modularity:} PROMPT\_EXPERT constrains each expert to output
    a \emph{single, concise, drop-in architectural cue} with explicit performance rationale.
    This prevents verbose conceptual drift and ensures each output corresponds to a manipulable
    graph-level transformation (consistent with $\mathcal{T}(a,i)$ in the main method).
    
    \item \textbf{Redundancy Control via Structured Merging:}
    PROMPT\_MERGE enforces JSON-based filtering and clustering, ensuring the orchestrator $\mathcal{M}$
    merges proposals into a \emph{minimal}, \emph{diverse}, and \emph{non-overlapping} inspiration set.
\end{itemize}


\paragraph{Effect on MMC Convergence.}
This prompt structure directly influences MMC stability:
\[
p_k^{(t)} = E_k(P, \mathcal{I}^{(t-1)}) \ \Rightarrow\ \mathcal{I}^{(t)} = \Omega(\{p_k^{(t)}\}).
\]
High-quality prompt specialization reduces noise in proposal trajectories, yielding:
$|\mu^{(t)}-\mu^{(t-1)}|$ decreases faster, and  $J(\mathcal{I}^{(t)},\mathcal{I}^{(t-1)})$ increases earlier.
Thus, convergence criteria described in Algorithm~\ref{alg:MMC} are met in fewer rounds while maintaining diversity.

\begin{algorithm}[t]
\caption{Multi-Round Multi-expert Consensus}
\label{alg:MMC}
\KwIn{Paper $P$; experts $\{E_1,\ldots,E_K\}$; thresholds $\tau_J,\tau_Q$;  round limits $t_{\min}, t_{\max}$};
\KwOut{Stable, diverse inspiration set $\mathcal{I}$}

$I^{(0)} \leftarrow \emptyset$, $t \leftarrow 1$\;

\While{not converged and $t < t_{\max}$}{
  \ForEach{$E_k$ in experts (in parallel)}{
    $p_k^{(t)} \leftarrow E_k(P, \mathcal{I}^{(t-1)})$\;
  }
  $\mathcal{I}^{(t)} \leftarrow \Omega(\{p_k^{(t)}\}_{k=1}^{K})$\;

  \BlankLine
  \tcp{Consensus stability based on set overlap}
  $J \leftarrow \frac{|\mathcal{I}^{(t)} \cap \mathcal{I}^{(t-1)}|}{|\mathcal{I}^{(t)} \cup \mathcal{I}^{(t-1)}|}$\;

  \BlankLine
  \tcp{Mean quality score based on reflection rewards; $r(i)$ is the reflection reward of inspiration $i$}
  $\mu^{(t)} \leftarrow \frac{1}{|\mathcal{I}^{(t)}|} \sum_{i \in \mathcal{I}^{(t)}} r(i)$\;

  $\Delta_\mu \leftarrow \frac{|\mu^{(t)}-\mu^{(t-1)}|}{|\mu^{(t-1)}| + \epsilon}$\;

  \If{$J \ge \tau_J$ \textbf{and} $\Delta_\mu \le \tau_Q$ \textbf{and} $t \ge t_{\min}$}{
    \textit{converged} $\leftarrow$ True\;
  }
  $t \leftarrow t + 1$\;
}
\Return{$\mathcal{I}^{(t-1)}$}\;
\end{algorithm}

\paragraph{Representation of Inspirations.}
Each inspiration $i\!\in\! \mathcal{I}_t$ is represented as a tuple
\[
i \equiv \langle \text{text}, \ \phi(i), \ u(i)\rangle,
\]
where \texttt{text} is a short, human-interpretable token sequence (\eg, \texttt{DWConv→SE→MLP}), $\phi(i)\in\mathbb{R}^d$ is a sentence/phrase embedding (\eg, via a frozen encoder) used for similarity/diversity computations, and $u(i)\in\mathbb{R}$ is a running utility updated based on credit from successful models. The mapping from $i$ to a concrete transformation $\mathcal{T}(\cdot,i)$ is defined by an \emph{operator template}.

\paragraph{Expert Decomposition.}
Given a paper $P$ with $K$ conceptual axes $S=\{s_1,\dots,s_K\}$ (\eg, hierarchy, receptive-field control, skip topology, gating/attention, normalization/activation, regularization, token routing), we instantiate experts $\{E_k\}_{k=1}^K$. Each expert specializes in one axis and proposes axis-specific cues conditioned on the current shared set $\mathcal{I}^{(t-1)}$:
\begin{equation}
p_k^{(t)} = E_k\big(P, \mathcal{I}^{(t-1)}\big),    
\end{equation}
yielding a multiset of proposals $\{p_k^{(t)}\}_{k=1}^K$. Proposals are short, “drop-in” primitives that can be composed in the block-graph grammar.

\paragraph{Consensus Operator.}
We use an orchestrator $\Omega$ to merge, filter, and deduplicate proposals:

\begin{equation}
    \mathcal{I}^{(t)} = \Omega\big(\{p_k^{(t)}\}_{k=1}^K\big).
\end{equation}
$\Omega$ performs lexical and semantic consolidation of proposals (implemented via an LLM merge prompt), while our outer loop uses sentence embeddings for stability and diversity checks.

\paragraph{Stopping Criterion and Stability.}
We check both set-stability and quality-plateau:
\begin{equation}
    \small{J\big(\mathcal{I}^{(t)},\mathcal{I}^{(t-1)}\big)\ge \tau_J,\quad
        \Delta_\mu^{(t)} \triangleq \frac{|\mu^{(t)}-\mu^{(t-1)}|}{|\mu^{(t-1)}|+\epsilon} \le \tau_Q,}
\end{equation}
with a minimum round count $t_{\min}$ to avoid premature stopping. 
This is exactly reflected in Alg.~\ref{alg:MMC} of the main paper.

\paragraph{Reliability of Sub-axis Decomposition}
To assess the reliability of our MMC-based sub-task decomposition, we repeatedly run the sub-task extractor under five independent seeds, as shown in~\cref{tab:expert_stability}. 
For each paper, we collect the set of sub-axis labels (\texttt{sub\_tasks}) produced in each run, and then treat each distinct label as a sub-axis candidate. We focus on \emph{recurring} sub-axes that appear in at least two runs, and measure their stability as
\begin{equation}
\small{\mathrm{consistency}(\ell) \;=\; \frac{\#\{\text{runs where label } \ell \text{ appears}\}}{\#\{\text{runs for that paper}\}} \in [0,1].}
\end{equation}
As a result, we obtain an average consistency of 84.98\% with a standard deviation of 16.59\%. This indicates that, once a sub-axis emerges as a recurring design dimension, it is typically recovered in the vast majority of runs, with only moderate variability. In other words, most architectural axes identified by our system are not artifacts of a particular prompt realization, but persist as stable roles under independent re-samplings of the LLM. The remaining variance reflects a smaller subset of more ambiguous or task-specific axes that are naturally sensitive to prompting and seed choices, highlighting both the robustness and residual stochasticity of the decomposition process.

\begin{table}[t]
\centering
\resizebox{\linewidth}{!}{
\begin{tabular}{lcc}
\toprule
\textbf{Metric} & \textbf{Value} & \textbf{Interpretation} \\
\midrule
Sub-axis Consistency (\%) & \textbf{84.98} & Most sub-axis labels recur across runs \\
Standard Deviation (\%)   & \textbf{16.59} & Moderate variability for a subset of axes \\
\bottomrule
\end{tabular}}
\caption{Stability of sub-axis labels across multiple random seeds. Higher consistency and lower variability indicate a more reliable decomposition.}
\label{tab:expert_stability}
\end{table}

\paragraph{Complexity.}
Each round costs $O(K + |\mathcal{I}^{(t)}|^2)$ in the worst case due to pairwise similarity computations used for stability and diversity monitoring.
In practice, we cap the number of proposals per expert, so $|\mathcal{I}^{(t)}|$ remains small and the overhead is negligible in our experiments.

\subsection{Adaptive Reflective Design Exploration}
\label{app:arde}

\myparagraph{Reflection-based Q-value estimation.}
Let $\mathcal{R} = \{(i^{(j)}, r^{(j)})\}_{j=1}^{M}$ denote the reflection log,
where each entry records the inspiration identifier $i^{(j)}$ and its reward
$r^{(j)}$ computed as the performance gain over its parent:
\begin{equation}
    r^{(j)} \;=\; \mathrm{acc}^{(j)}_{\text{child}} - \mathrm{acc}^{(j)}_{\text{parent}}.
\end{equation}
For a given inspiration $i$, we collect the index set
\begin{equation}
    \mathcal{J}(i) \;=\; \{\, j \mid i^{(j)} = i \,\},
\end{equation}
and define its reflection-based Q-value as the empirical mean reward:
\begin{equation}
    Q(i)
    \;=\;
    \begin{cases}
        \dfrac{1}{|\mathcal{J}(i)|} \displaystyle\sum_{j \in \mathcal{J}(i)} r^{(j)},
        & \text{if } |\mathcal{J}(i)| > 0,\\[1.0em]
        0, & \text{otherwise.}
    \end{cases}
    \label{eq:q_from_reflection}
\end{equation}
These Q-values are then used to bias inspiration sampling in ARDE, 
favoring inspirations that consistently yield positive improvements.

\paragraph{Redundancy Filter.}
Before materialization, we apply redundancy pruning over proposals using semantic radius $\delta$ in embedding space:
\[
\text{Keep}\; i \ \text{only if}\ \min_{j\in \text{Kept}} d_{\cos}\big(\phi(i),\phi(j)\big) \ge \delta.
\]
We also drop proposals that violate hard feasibility checks derived from block-graph templates (e.g., incompatible channel sizes or stride cascades).

\paragraph{Exploration/Exploitation Gate.}
ARDE (Sec. 3.2) modulates the fraction of \emph{exploratory} proposals admitted. 
When reward variance falls, we accept a larger share of structurally novel or out-of-axis proposals; when variance spikes, we prioritize exploitative refinements around high-utility inspirations.

\paragraph{Inspiration Utility Update.}
After selection, inspirations that contributed to survivors receive positive credit; those that contributed to discarded candidates may receive neutral or slight decay:
\begin{equation}
u(i)\leftarrow \gamma \cdot u(i) + \sum_{m\in \mathcal{S}_t: i\in U(m)} \kappa \cdot R(m),
\quad \gamma\in(0,1],\ \kappa>0.    
\end{equation}
This utility $u(i)$ is used by the adaptive retriever $\mathcal{R}_t$ to bias sampling in the next generation (softmax over $u(i)$ with temperature).

\paragraph{Memory Aging and Curation.}
We age out inspirations with persistently low $u(i)$ to keep memory compact, and maintain axis-wise quotas to preserve conceptual coverage.

\subsection{Pareto-Guided Evolutionary Selection}
\label{app:pes}

\paragraph{Structural Diversity.}
We require a structural diversity signal between a child block $G_c$ and its parent $G_p$ that is (i) cheap to compute for hundreds of candidates and (ii) more stable than exact graph edit distance (GED), which is sensitive to small local rewires. To this end, we define a bounded structural similarity $S \in [0,1]$ and use
\begin{equation}
    \text{StructDiv.}(G_c, G_p) \;=\; 1 - S,
\end{equation}
so that higher values indicate more structurally novel children.
Specifically, each block is represented as a directed graph $G=(V,E)$ with node labels $\ell_G(v)$ denoting operation types. We combine three complementary similarities:
(1) Operation composition.
We first compare the multiset of operation types via a Jaccard similarity
\begin{equation}
    h_G(o) = \big|\{ v \in V_G \mid \ell_G(v)=o \}\big|,
\end{equation}
\begin{equation}
    s_{\mathrm{op}} =
    \frac{\sum_o \min\!\big(h_c(o), h_p(o)\big)}
         {\sum_o \max\!\big(h_c(o), h_p(o)\big)},
\end{equation}
with the convention $s_{\mathrm{op}}=1$ if the denominator is zero. This term checks whether parent and child use a similar mix of building blocks.
(2) Depth-aware placement.
Then, we assign each node a topological depth $d_G(v)$ and build, for each operation $o$, a depth histogram $c_G^{(o)}(k)$ and normalized distribution $p_G^{(o)}(k)$. For each $o$, we compute a 1D Wasserstein-1 distance along the depth axis via cumulative distributions and convert it to a similarity $s_{\mathrm{depth}}^{(o)} = 1 - \mathrm{EMD}^{(o)}$. Averaging over all operations that appear in either graph gives
\begin{equation}
    s_{\mathrm{depth}} = \frac{1}{|\mathcal{O}'|} \sum_{o \in \mathcal{O}'} s_{\mathrm{depth}}^{(o)},
\end{equation}
which captures whether the same operations are placed at similar stages of the computation.
(3) Local wiring motifs.
Finally, we run a few iterations ($T{=}2$) of Weisfeiler--Lehman color refinement, starting from operation labels, and build a color histogram $h_G^{\mathrm{WL}}$ at the final iteration. We then compute the cosine similarity
\begin{equation}
    s_{\mathrm{WL}} \;=\;
    \frac{\sum_{\gamma} h_c^{\mathrm{WL}}(\gamma) h_p^{\mathrm{WL}}(\gamma)}
         {\sqrt{\sum_{\gamma} h_c^{\mathrm{WL}}(\gamma)^2}\;
          \sqrt{\sum_{\gamma} h_p^{\mathrm{WL}}(\gamma)^2}},
\end{equation}
which provides a lightweight proxy for higher-order wiring patterns.
Here, the three terms are combined as
\begin{equation}
    S_{\mathrm{raw}} 
    = 0.50\, s_{\mathrm{op}}
    + 0.30\, s_{\mathrm{depth}}
    + 0.20\, s_{\mathrm{WL}},
\end{equation}
\begin{equation}
    S = \min\big(1,\; \max(0, S_{\mathrm{raw}})\big),
\end{equation}
and
\begin{equation}
    \text{StructDiv.}(G_c, G_p) =
    \begin{cases}
        1, & \text{if } G_c \text{ or } G_p \text{ is missing},\\[3pt]
        1 - S, & \text{otherwise.}
    \end{cases}
\end{equation}
This factorized design is significantly cheaper than full GED, yet still aligns with architectural intuition: novel children should differ in which operations they use, where they place them, and how they wire them into local motifs.

\begin{table}[t]
    \centering
    \begin{subtable}[t]{\linewidth}
        \centering
        \resizebox{\linewidth}{!}{%
        \begin{tabular}{ccc|cccc}
        \toprule
         \multicolumn{3}{c|}{\textbf{Pareto}} & \multicolumn{4}{c}{\textbf{CIFAR10}}  \\
         \midrule
         \textbf{Params $\lambda_p$} & \textbf{Conf. $\rho_c$} & \textbf{Latency $\lambda_{l}$} & \textbf{\# Params} & \textbf{GFLOPs} & \textbf{Latency} & \textbf{Acc.}\\
         \midrule
         0.10 & 0.10 & 0.10 
         & 1.38M & 0.200 & 9.3ms & 94.45 \\
         0.25 & 0.10 & 0.10 
         & 1.28M & 0.185 & 10.7ms & 94.41 \\
         0.10 & 0.25 & 0.10 
         & 1.38M & 0.200 & 6.0ms & 94.76  \\
         0.10 & 0.10 & 0.25
         & 1.38M & 0.200 & 9.0ms & 94.49 \\
        \bottomrule
        \end{tabular}}
        \caption{CIFAR10.}
    \end{subtable}

    \vspace{2pt}

    \begin{subtable}[t]{\linewidth}
        \centering
        \resizebox{\linewidth}{!}{%
        \begin{tabular}{ccc|cccc}
        \toprule
         \multicolumn{3}{c|}{\textbf{Pareto}} & \multicolumn{4}{c}{\textbf{ImageNet16-120}}  \\
         \midrule
         \textbf{Params $\lambda_p$} & \textbf{Conf. $\rho_c$} & \textbf{Latency $\lambda_{l}$} & \textbf{\# Params} & \textbf{GFLOPs} & \textbf{Latency} & \textbf{Acc.}\\
         \midrule
         0.10 & 0.10 & 0.10 
         & 1.26M & 0.049 & 9.1ms & 50.22 \\
         0.25 & 0.10 & 0.10 
         & 1.16M & 0.049 & 9.0ms & 48.88 \\
         0.10 & 0.25 & 0.10 
         & 1.26M & 0.049 & 14.7ms & 49.51 \\
         0.10 & 0.10 & 0.25
         & 1.26M & 0.049 & 7.2ms & 49.22 \\
        \bottomrule
        \end{tabular}}
        \caption{ImageNet16-120.}
    \end{subtable}
    \caption{Effect of Pareto preference weights on architecture selection.}
    \label{tab:pareto}
\end{table}

\paragraph{Confidence.}
Given a trained classifier $f_\theta$ and an evaluation set
$\mathcal{D} = \{x_i\}_{i=1}^N$, we define the predictive probability
vector for each input as
\begin{equation}
    \mathbf{p}_i \;=\; \mathrm{softmax}\!\big(f_\theta(x_i)\big)
    \in \mathbb{R}^C,
\end{equation}
where $C$ is the number of classes.
The per-sample confidence is the maximum class probability
\begin{equation}
    c_i \;=\; \max_{1 \leq k \leq C} \, p_{i,k}.
\end{equation}
The reported \emph{robustness} score corresponds exactly to the average
confidence over all test samples:
\begin{equation}
    \text{Conf.}
    \;=\;
    \frac{1}{N} \sum_{i=1}^{N} c_i
    \;=\;
    \frac{1}{N} \sum_{i=1}^{N}
    \max_{k} \, \mathrm{softmax}\!\big(f_\theta(x_i)\big)_k.
\end{equation}

\subsection{Hyperparameters and Sensitivity (Defaults)}
\begin{itemize}
\item \textbf{MMC:} $\tau_J\in[0.5,0.8]$, $\tau_Q\in[0.01,0.05]$, $t_{\min}=2$, $t_{\max}=5\sim 8$.
\item \textbf{ARDE:} $\varepsilon_{\min}=0.05$, $\varepsilon_{\max}=0.5$, $\lambda\in[1,5]$, window $k=3 \sim 10$ events.
\item \textbf{PES:} survival ratio $r=0.3\sim 0.6$; $\lambda_p,\lambda_l,\gamma_d,\rho_c,\beta$ chosen so that each normalized term is $\mathcal{O}(1)$ and no single dimension dominates (we use grid-tuned presets).
\end{itemize}

\begin{figure}
    \centering
    \includegraphics[width=1\linewidth]{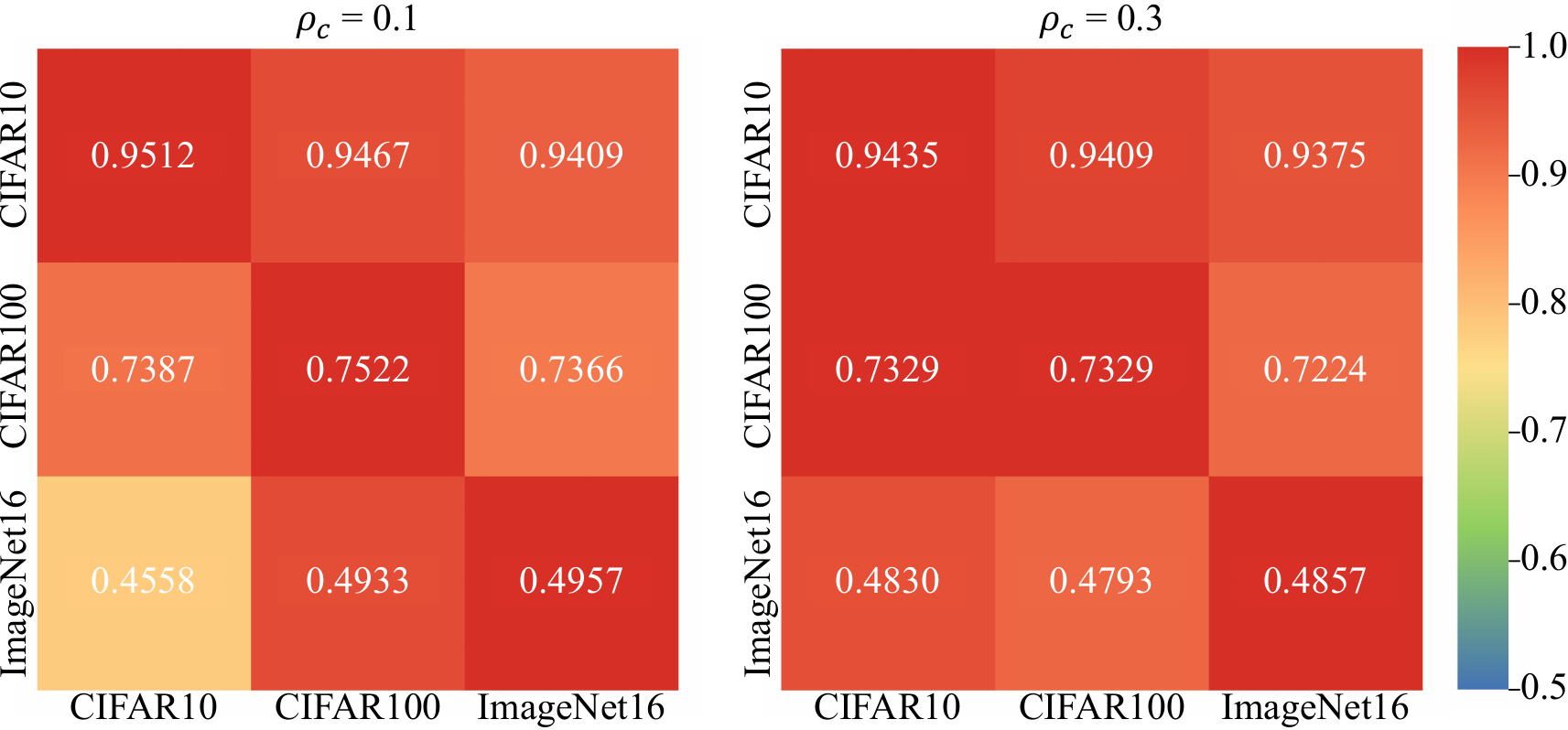}
    \caption{Cross-dataset direct transfer under different confidence weights $\rho_c$.
    Each heatmap reports the test accuracy, with colors indicating the accuracy deviation, when architectures searched on a \emph{source} dataset (rows) are directly transferred and evaluated on a \emph{target} dataset (columns). Diagonal entries denotes \emph{Oracle} in-domain performance where the source and target datasets are the same.}
    \label{fig:conf_transfer}
\end{figure}

\section{Additional Experiments}
\label{app:exp}

\subsection{Effect of Pareto Preference Weights}

\cref{tab:pareto} analyzes how the Pareto preference weights $(\lambda_p,\rho_c,\lambda_l)$ affect the final architectures on CIFAR10 and ImageNet16-120. 
We start from a balanced setting $(0.10,0.10,0.10)$ and then increase each axis in turn while keeping the others fixed.
On CIFAR10, increasing the parameter penalty to $\lambda_p=0.25$ produces a slightly smaller model (1.28M vs.\ 1.38M parameters and 0.185 vs.\ 0.200 GFLOPs) with trivial accuracy change (94.41\% vs.\ 94.45\%), showing that RevoNAD can trade a modest amount of capacity for compactness without sacrificing performance. 
Emphasizing confidence with $\rho_c=0.25$ keeps the model size roughly fixed but yields the best accuracy (94.76\%) and the lowest latency (6.0\,ms), indicating that the confidence-aware term guides the search toward architectures that are both reliable and efficient. 
When latency is upweighted ($\lambda_l=0.25$), the selected architecture attains comparable accuracy (94.49\%) with similar complexity, confirming that the controller can prioritize runtime without collapsing performance.
On ImageNet16-120, the trends are similar but the trade-offs are more pronounced. 
A stronger parameter penalty ($\lambda_p=0.25$) reduces the model to 1.16M parameters while incurring a moderate drop in accuracy (50.22\% $\rightarrow$ 48.88\%). 
Upweighting confidence or latency mainly shifts the operating point along the accuracy–latency frontier: $\rho_c=0.25$ leads to a slower but slightly less accurate model (14.7\,ms, 49.51\%), whereas $\lambda_l=0.25$ yields the fastest configuration (7.2\,ms) with a small degradation to 49.22\%. 
Overall, these results indicate that the Pareto-weighted scoring in RevoNAD is stable and interpretable, allowing the search procedure to be systematically tuned toward compact or low-latency architectures while maintaining competitive accuracy.

\subsection{Architectural Robustness}

We further examine the generalization ability of RevoNAD by directly transferring architectures across datasets while varying the confidence weight $\rho_c$ in the Pareto objective.
For each setting of $\rho_c$, we first run the search pipeline on a \emph{source} dataset (CIFAR10, CIFAR100, or ImageNet16-120), and then evaluate the discovered architectures on all three datasets without any additional search or fine-tuning.
The resulting cross-dataset accuracies are summarized as heatmaps in Fig.~\ref{fig:conf_transfer}.
First, the diagonal entries remain high across both $\rho_c=0.1$ and $\rho_c=0.3$, indicating that the architectures found by RevoNAD are strong on their respective source tasks (\ie, around $95\%$ on CIFAR10 and mid-$70\%$ on CIFAR100), regardless of the exact confidence weighting.
Here, we observe that increasing $\rho_c$ from $0.1$ to $0.3$ contributes to address the domain shift from ImageNet16-120 to CIFAR10, recovering up to +2.71\% accuracy. 
Overall, these results indicate that RevoNAD exhibits stable cross-dataset generalization leveraging the confidence weight, and that architectures discovered on a single source benchmark can be reliably reused on related image classification tasks.

\begin{table}[t]
    \begin{center}
    \resizebox{\linewidth}{!}{%
    \begin{tabular}{l|C{1.2cm}C{1.2cm}|C{1.2cm}C{1.2cm}|C{1.2cm}C{1.2cm}}
    \toprule
 \multirow{2.3}{*}{\textbf{Model}} & \multicolumn{2}{c|}{\textbf{CIFAR10}~($\uparrow$)}   & \multicolumn{2}{c|}{\textbf{CIFAR100}~($\uparrow$)}  & \multicolumn{2}{c}{\textbf{ImageNet16-120}~($\uparrow$)} \\
 \cmidrule{2-7}
 & \textbf{Val.} & \textbf{Test} & \textbf{Val.} & \textbf{Test} & \textbf{Val.} & \textbf{Test} \\
 \midrule
 DeepSeek
 & 91.28 & 94.49 & 75.18 & 74.83 & 48.87 & 49.13 \\
 Qwen2.5
& 91.17 & 94.77 & 76.21 & 76.25 & \textbf{51.00} & 50.55 \\
GPT-4o
& \textbf{92.55} & \textbf{95.22} & \textbf{76.70} & \textbf{76.38} & 50.58 & \textbf{50.72} \\
\bottomrule
\end{tabular}}
\end{center}
\caption{Effect of the underlying LLM on our framework.}
\label{tab:llm}
\end{table}

\begin{table*}[t]
\centering
\renewcommand{\arraystretch}{1.15}
\resizebox{\linewidth}{!}{%
\begin{tabular}{c c c p{0.8\linewidth}}
\toprule
\textbf{Inspiration ID} & \textbf{Dataset} & \textbf{Acc. (\%)} & \textbf{Inspiration} \\

\midrule
3161 & CIFAR10 & 0.0 & per-block top-k expert router (MoE-lite) with load-balance loss \\
7986 & CIFAR10 & 0.0 & reprojection-based feature alignment \\
13601 & CIFAR10 & 0.0 & Design with passive solar orientation to enhance energy efficiency and comfort. \\
\midrule
10039 & CIFAR100 & 1.0 & in-block cross-scale fusion (stage k2194 $\leftrightarrow$ k00b11) via 1x1 proj + add \\
10083 & CIFAR100 & 1.0 & anti-alias downsample branch (blur$\rightarrow$stride-2 conv) within block \\
12206 & CIFAR100 & 1.0 & RMSNorm swap-in + pre-norm residual; no projection when dims equal \\
\midrule
38894 & ImageNet16-120 & 0.83 & Adaptive reuse of historic buildings with modular interiors for modern functionality and sustainability. \\
21051 & ImageNet16-120 & 6.37 & Biophilic skyscrapers integrating vertical gardens for sustainable urban living. \\
40959 & ImageNet16-120 & 13.43 & Solar-powered community centers with adaptive spaces for diverse cultural events. \\
\bottomrule
\end{tabular}}
\vspace{-1em}
\caption{Examples of inspiration collapse across datasets.}
\label{tab:axis_collapse}
\end{table*}

\begin{table*}[h!]
\centering
\resizebox{\linewidth}{!}{%
\begin{tabular}{l}
\toprule
\textbf{Error logs} \\
\midrule
\texttt{node 1 error: Output shape must be (B,dim,H,W).} \\
\texttt{node 23 error: reshape operation's first dimension must be B.} \\
\texttt{node 21 error: The node is not used.} \\
\texttt{node 7 error: The node is not used.} \\
\texttt{node 19 error: Add operation's inputs can not be Added.} \\
\texttt{node 8 error: Undefined computation max\_pool2d is used.} \\
\texttt{node 10 error: When using Linear, the last dimension must be the channel dimension.} \\
\texttt{node 6 error: Conv2d operation can receive only one input.} \\
\texttt{node {node} error: H and W cannot appear in layer definition.} \\
\texttt{shape '[4, 256]' is invalid for input of size 2048.} \\
\texttt{node 1 error: The height and width of the output feature map must be 1.0 of the input feature map.} \\
\texttt{node 19 error: repeat operation can receive only one input.} \\
\texttt{output must be the only output node.} \\
\texttt{The computation graph of the block is not a directed acyclic graph.} \\
\bottomrule
\end{tabular}}
\vspace{-1em}
\caption{Failure cases during architecture developing.}
\label{tab:develop_error}
\end{table*}

\subsection{Structural Design Diversity}

\begin{figure*}[h!]
    \centering
    \includegraphics[width=0.8\linewidth]{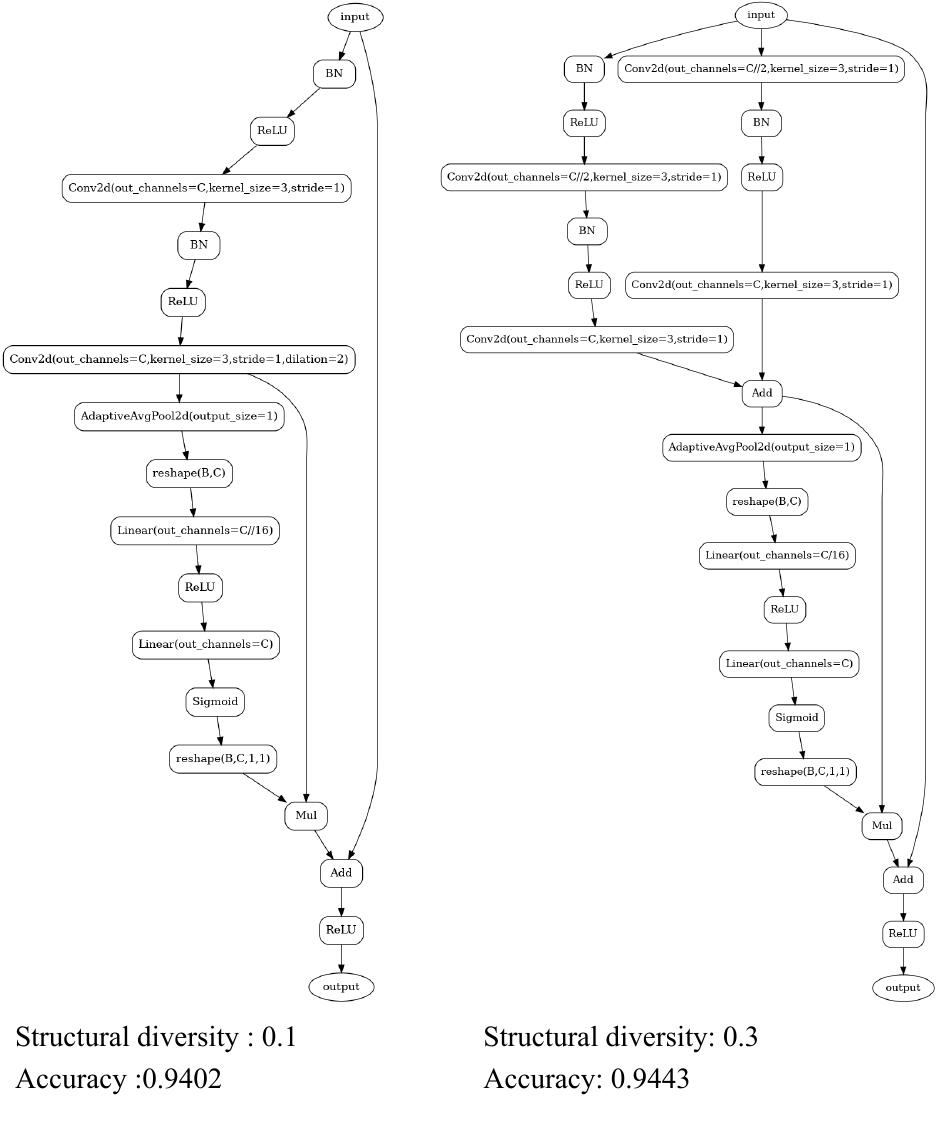}
    \caption{Structural Diversity on CIFAR10}
    \label{fig:structdiv_cifar10}
\end{figure*}

\begin{figure*}[h!]
    \centering
    \includegraphics[width=0.8\linewidth]{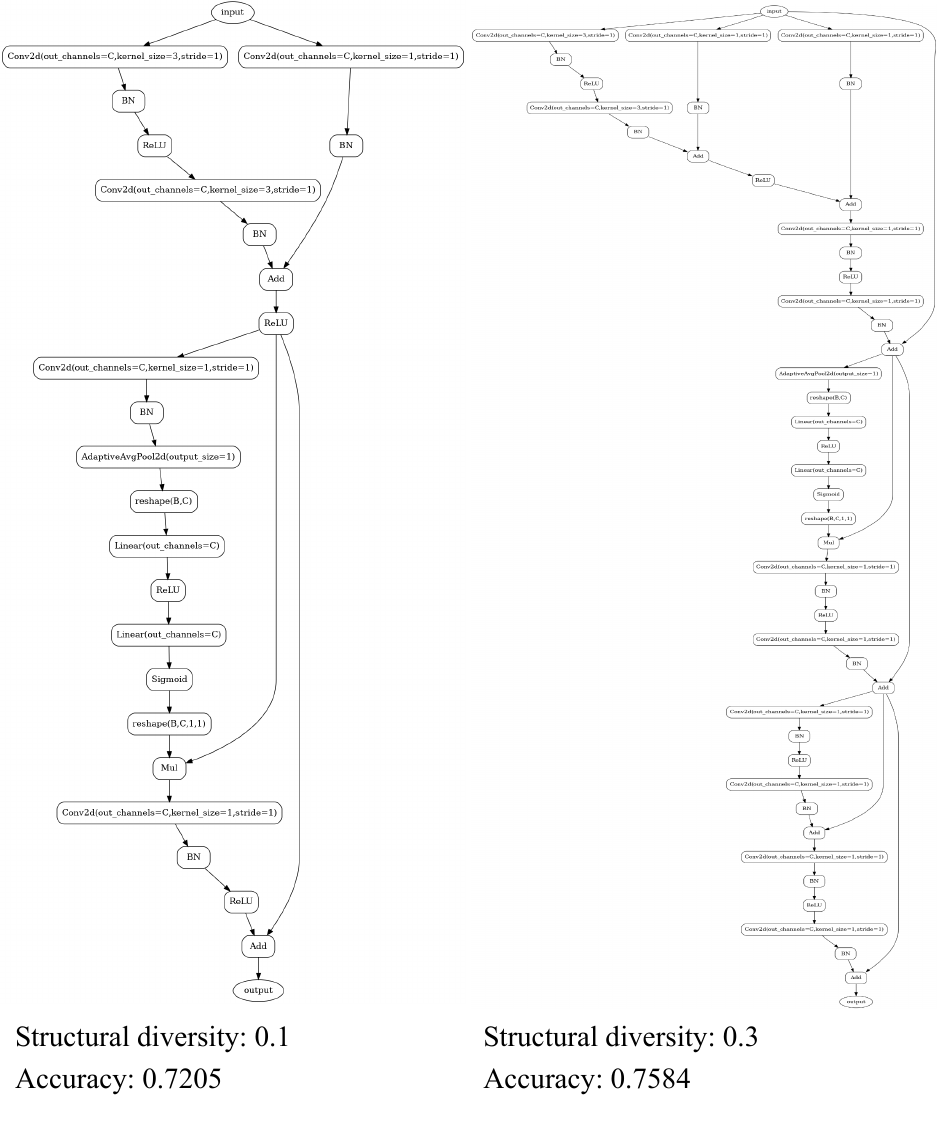}
    \caption{Structural Diversity on CIFAR100}
    \label{fig:structdiv_cifar100}
\end{figure*}

\begin{figure*}[h!]
    \centering
    \includegraphics[width=0.8\linewidth]{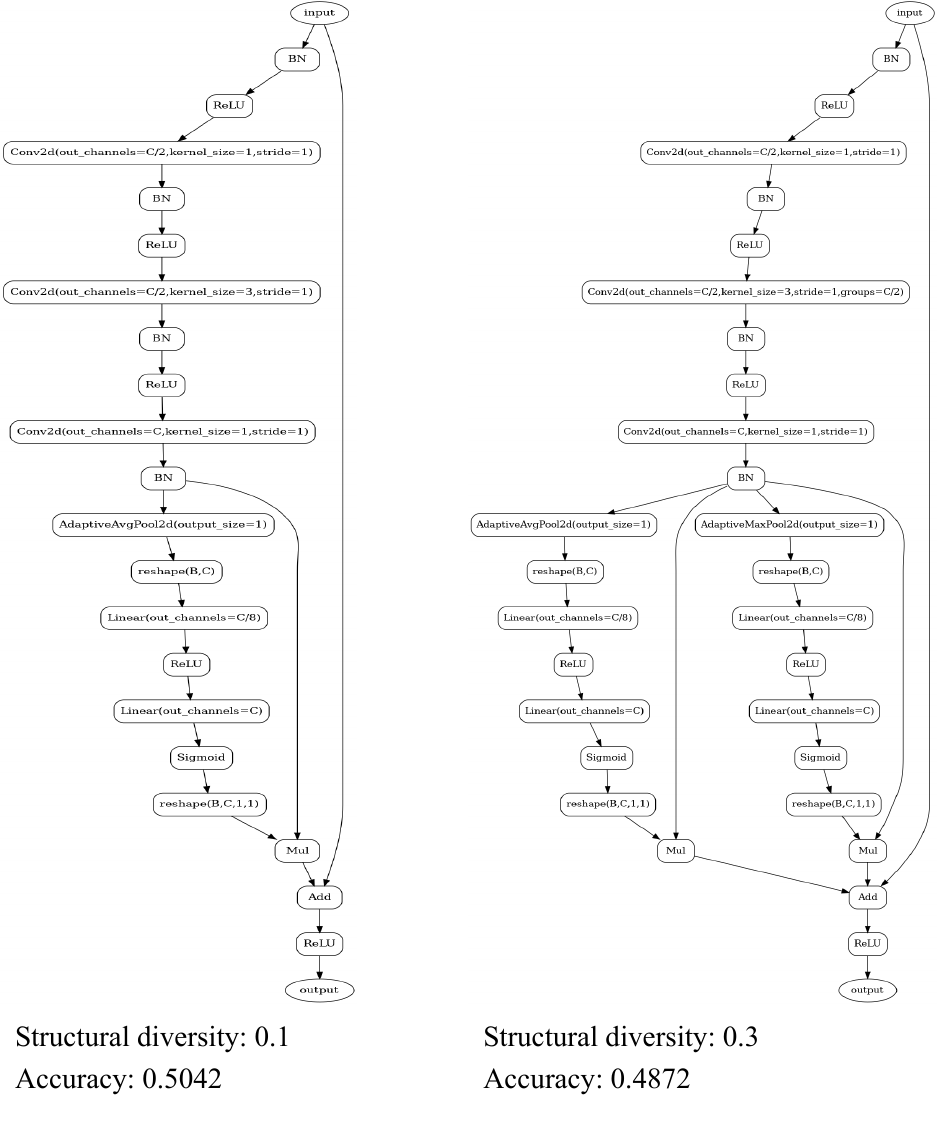}
    \caption{Structural Diversity on ImageNet16-120}
    \label{fig:structdiv_imagenet}
\end{figure*}

We study how the structural-diversity weight $\gamma_d$ in our PES influences the discovered blocks. For each dataset (CIFAR10, CIFAR100, and ImageNet16-120), we run RevoNAD with $\gamma_d \in \{0.1, 0.3\}$ while keeping all other hyperparameters fixed and visualize the resulting best blocks.
On CIFAR10~\cref{fig:structdiv_cifar10}, increasing the diversity weight from $0.1$ to $0.3$ produces a slightly richer topology with additional residual paths and aggregation operations, which translates into a modest but consistent accuracy gain (from 94.02\% to 94.43\%). 
The effect is more pronounced on CIFAR100~\cref{fig:structdiv_cifar100}. With $\gamma_d = 0.3$, the discovered block exhibits deeper stacking and more heterogeneous branches compared to the relatively shallow pattern obtained with $\gamma_d = 0.1$, leading to a substantial improvement from 72.05\% to 75.84\% accuracy. 
In contrast, on ImageNet16-120~\cref{fig:structdiv_imagenet}, a stronger emphasis on diversity slightly degrades performance: the model found with $\gamma_d = 0.3$ is structurally more fragmented, but its accuracy drops from 50.42\% to 48.72\%. 
Overall, it is noteworthy that a moderate increase in $\gamma_d$ yields clear gains on CIFAR10/100 by promoting new architectural motifs, whereas on ImageNet16-120 an overly strong diversity prior can harm performance, suggesting that $\gamma_d$ should be tuned to balance novelty and reliability.

\subsection{Effect of the Underlying LLM}
\cref{tab:llm} reports the impact of different underlying LLMs on our framework.
On CIFAR10, GPT-4o achieves the best validation and test accuracies (92.55$\%$ / 95.22$\%$), outperforming DeepSeek (91.28$\%$ / 94.49$\%$) and Qwen2.5 (91.17$\%$ / 94.77$\%$).
On CIFAR100, the differences between GPT-4o and Qwen2.5 become smaller: GPT-4o reaches 76.70$\%$ / 76.38$\%$ (val/test), while Qwen2.5 attains 76.21$\%$ / 76.25$\%$.
On the more challenging ImageNet16-120 benchmark, performance across LLMs is largely comparable.
Qwen2.5 obtains the highest validation accuracy (51.00$\%$), whereas GPT-4o delivers the best test accuracy (50.72$\%$), with all gaps within 0.6 points.
These results indicate that our framework is not overly sensitive to the particular LLM instantiation: stronger models such as GPT-4o provide consistent but moderate gains, while more lightweight alternatives (DeepSeek, Qwen2.5) still produce competitive architectures without collapsing the search.

\subsection{Best-performing models (95.22\%)}
As shown in~\cref{tab:llm_explore_examples_full}, we present the best-performing architectural lineage discovered by our framework and summarize the key architectural modifications for each model.
\begin{itemize}
  \item \textbf{baseline} (90.80\%): ResNet basic block.
  \item \textbf{p8515} (95.22\%): Replace 2nd conv by DWConv $\rightarrow$ PWConv path; extra PWConv and residual add for spatial token mixing.
  \item \textbf{p1483} (95.22\%): DW-FFN base; add block-wise routing/gating idea (MoE-lite) while keeping residual PW mixing.
  \item \textbf{p9167} (95.22\%): DWConv(3$\times$3, groups=C) $\rightarrow$ PWConv $\rightarrow$ PWConv with residual add; token mixing in FFN style.
\end{itemize}

\subsection{Failure Modes}

\paragraph{Inspiration Collapse.}
As shown in~\cref{tab:axis_collapse}, RevoNAD occasionally produces infeasible or non-viable architectures for two main reasons.
First, although our MMC module generates structurally meaningful inspirations via iterative collaborative discussions, the underlying LLM can still hallucinate structural design motifs.
Second, even though the inspiration memory is populated with semantically diverse design cues, a relatively high exploration rate $\varepsilon_n$ can steer the search toward mismatched or non-informative architectural clues, which are more likely to lead to invalid or degenerate architectures.
Due to these limitations, bridging architectural search with LLM-driven generation remains a challenging problem.

\paragraph{Developing failures.}
NAD effectively addresses the narrow, hand-crafted search spaces of conventional NAS and can propose well-informed designs grounded in high-level architectural knowledge. 
However, NAD still faces difficulties in consistently aligning high-level specifications with low-level operator modules, mainly due to detailed code implementation error. 
As illustrated by the error logs in Table~\ref{tab:develop_error}, the generated blocks suffer from implementation-time failures such as invalid output shapes, mismatched batch or channel dimensions, illegal multi-input usage of operators (e.g., \texttt{Conv2d} or \texttt{repeat}), undefined operations (\eg, \texttt{max\_pool2d}), unused or multiply defined nodes, and computation graphs that are not directed acyclic graphs. 
In this section, we present a practical limitation of our pipeline: a non-trivial portion of the search budget is spent on architectures that appear syntactically plausible at the design level but cannot be realized as valid, trainable models.

\begin{table*}[h!]
\centering
\small
\renewcommand{\arraystretch}{1.15}
\begin{tabular}{l p{5.2cm} c p{7.5cm}}
\toprule
\textbf{Inspiration ID} & \textbf{Inspiration} & \textbf{Acc. (\%)} & \textbf{Key Architectural Modification} \\
\midrule
baseline & ResNet basic block & 90.80 & Two 3$\times$3 conv + BN + ReLU with residual add (standard baseline). \\
p8515 & DW-FFN: depthwise conv inside MLP for token mixing & 95.22 & Replace 2nd conv by DWConv$\rightarrow$PWConv path; extra PWConv and residual add for spatial token mixing. \\
p1483 & Per-block top-$k$ expert router (MoE-lite) with load-balance loss & 95.22 & DW-FFN base; add block-wise routing/gating idea (MoE-lite) while keeping residual PW mixing. \\
p9167 & DW-FFN: depthwise conv inside MLP for token mixing & 95.22 & DWConv($3\times3$, groups=$C$)$\rightarrow$PWConv$\rightarrow$PWConv with residual add; token mixing in FFN style. \\
p965 & DW-FFN: depthwise conv inside MLP for token mixing & 95.12 & Compact DWConv($3\times3$)$\rightarrow$PWConv; remove last ReLU; pre-activation-style add. \\
p477 & DW-FFN: depthwise conv inside MLP for token mixing & 95.12 & DWConv($3\times3$)$\rightarrow$PWConv with BN then residual add; lightweight token mixer. \\
p12347 & RMSNorm swap-in + pre-norm residual & 95.12 & Pre-norm variant of DW-FFN idea (implemented via BN here) before residual add for stability. \\
p8334 & DWConv($k{=}5$)$\rightarrow$PWConv expand & 94.98 & Replace inner 3$\times$3 with DWConv($5$) then PWConv expansion; residual add kept minimal. \\
p2571 & DW-FFN: depthwise conv inside MLP for token mixing & 94.94 & Minimal DWConv($3$)$\rightarrow$PWConv pathway attached to residual. \\
p6793 & DWConv($k{=}7$)$\rightarrow$PWConv expand 4$\times$ (MBConv variant) & 94.92 & Wider DW kernel ($7$) then PW expand; MBConv-like expansion under residual. \\
p8158 & Task-agnostic expert learning via gating re-param & 94.87 & Output-side GELU gating on residual to encourage expert-like specialization. \\
p7852 & DWConv($k{=}7$)$\rightarrow$PWConv expand 4$\times$ (MBConv variant) & 94.86 & Stronger MBConv-style expansion (4$\times$) inside residual token mixer. \\
p9908 & RMSNorm swap-in + pre-norm residual & 94.70 & Pre-norm (via BN) after residual add to stabilize gradients. \\
p1865 & In-block cross-scale fusion (stage $k\leftrightarrow k{\pm}1$) & 94.69 & Extra 1$\times$1 projection branch merged into residual to blend cross-scale features. \\
p7257 & Anti-alias downsample branch (blur/dilation) within block & 94.68 & Use dilation in DW-FFN path to reduce aliasing while keeping residual mixing. \\
p2455 & RMSNorm swap-in + pre-norm residual & 94.67 & BN-as-pre-norm placed on residual output; remove extra nonlinearity. \\
p4569 & RMSNorm swap-in + pre-norm residual & 94.64 & Pre-norm BN at input; output-side BN after add. \\
p4199 & Cross-token routing head (top-$k$ sparse links) & 94.61 & Add light routing on DW-FFN output path to sparsify cross-token flow. \\
p928 & Use pre-norm residual connections for block stability & 94.57 & BN$\rightarrow$Conv$\rightarrow$ReLU pre-activation; residual add before final BN. \\
p11876 & Dynamic routing head for enhanced feature selection & 94.55 & DW-FFN with dilated DWConv($d{=}2$) plus selection head at output. \\
p3128 & RMSNorm swap-in + pre-norm residual & 94.52 & Extra normalization after residual add for stability under small models. \\
p11944 & DW-FFN: depthwise conv inside MLP for token mixing & 94.50 & Minimal DWConv($3$)$\rightarrow$PWConv with straight residual add (no extra BN). \\
p351 & RMSNorm swap-in + pre-norm residual & 94.49 & Remove tail ReLU; pre-activation residual with BN to improve stability. \\
p8417 & In-block cross-scale fusion via 1$\times$1 proj + add & 94.43 & Add 1$\times$1 projection and fuse into residual to mix neighbor-scale features. \\
p9874 & DW-FFN: depthwise conv inside MLP for token mixing & 94.31 & DWConv($3$)$\rightarrow$PWConv$\rightarrow$PWConv chain before add. \\
p4616 & DW-FFN variant & 94.25 & Second conv becomes DWConv; extra 1$\times$1 to align channels before add. \\
p8060 & Upconv with residual scaling & 94.22 & Dilated DW path + residual rescaling via extra conv then activation. \\
p856 & Anti-alias downsample branch (blur$\sim$dilation) & 94.17 & Replace stride-2 with dilation on 2nd conv to suppress aliasing in residual path. \\
\bottomrule
\end{tabular}
\caption{Architectures generated by the auto-reflective $\varepsilon$-greedy explorer starting from a ResNet basic block (CIFAR10). We list the strongest and diverse variants (omitting degenerate 0/10\% runs).}
\label{tab:llm_explore_examples_full}
\end{table*}

%% file: main.bbl
\begin{thebibliography}{65}
\providecommand{\natexlab}[1]{#1}
\providecommand{\url}[1]{\texttt{#1}}
\expandafter\ifx\csname urlstyle\endcsname\relax
  \providecommand{\doi}[1]{doi: #1}\else
  \providecommand{\doi}{doi: \begingroup \urlstyle{rm}\Url}\fi

\bibitem[Achiam et~al.(2023)Achiam, Adler, Agarwal, Ahmad, Akkaya, Aleman, Almeida, Altenschmidt, Altman, Anadkat, et~al.]{achiam2023gpt}
Josh Achiam, Steven Adler, Sandhini Agarwal, Lama Ahmad, Ilge Akkaya, Florencia~Leoni Aleman, Diogo Almeida, Janko Altenschmidt, Sam Altman, Shyamal Anadkat, et~al.
\newblock Gpt-4 technical report.
\newblock \emph{arXiv preprint arXiv:2303.08774}, 2023.

\bibitem[Bai et~al.(2023)Bai, Bai, Chu, Cui, Dang, Deng, Fan, Ge, Han, Huang, et~al.]{bai2023qwen}
Jinze Bai, Shuai Bai, Yunfei Chu, Zeyu Cui, Kai Dang, Xiaodong Deng, Yang Fan, Wenbin Ge, Yu Han, Fei Huang, et~al.
\newblock Qwen technical report.
\newblock \emph{arXiv preprint arXiv:2309.16609}, 2023.

\bibitem[Bergstra and Bengio(2012)]{bergstra2012random}
James Bergstra and Yoshua Bengio.
\newblock Random search for hyper-parameter optimization.
\newblock \emph{The journal of machine learning research}, 13\penalty0 (1):\penalty0 281--305, 2012.

\bibitem[Bolya et~al.(2019)Bolya, Zhou, Xiao, and Lee]{bolya2019yolact}
Daniel Bolya, Chong Zhou, Fanyi Xiao, and Yong~Jae Lee.
\newblock Yolact: Real-time instance segmentation.
\newblock In \emph{Proceedings of the IEEE/CVF international conference on computer vision}, pages 9157--9166, 2019.

\bibitem[Chen et~al.(2023)Chen, Dohan, and So]{chen2023evoprompting}
Angelica Chen, David Dohan, and David So.
\newblock Evoprompting: Language models for code-level neural architecture search.
\newblock \emph{Advances in neural information processing systems}, 36:\penalty0 7787--7817, 2023.

\bibitem[Chen et~al.(2025)Chen, Du, Dai, Zhang, Wang, Wang, Tang, Zhang, and Yu]{chen2025debatecoder}
Jizheng Chen, Kounianhua Du, Xinyi Dai, Weiming Zhang, Xihuai Wang, Yasheng Wang, Ruiming Tang, Weinan Zhang, and Yong Yu.
\newblock Debatecoder: Towards collective intelligence of llms via test case driven llm debate for code generation.
\newblock In \emph{Proceedings of the 63rd Annual Meeting of the Association for Computational Linguistics (Volume 1: Long Papers)}, pages 12055--12065, 2025.

\bibitem[Chen et~al.(2024)Chen, Han, and Zhang]{chen2024comm}
Pei Chen, Boran Han, and Shuai Zhang.
\newblock Comm: Collaborative multi-agent, multi-reasoning-path prompting for complex problem solving.
\newblock \emph{arXiv preprint arXiv:2404.17729}, 2024.

\bibitem[Chen et~al.(2020)Chen, Wang, Cheng, Tang, and Hsieh]{chen2020drnas}
Xiangning Chen, Ruochen Wang, Minhao Cheng, Xiaocheng Tang, and Cho-Jui Hsieh.
\newblock Drnas: Dirichlet neural architecture search.
\newblock \emph{arXiv preprint arXiv:2006.10355}, 2020.

\bibitem[Choi et~al.(2025)Choi, Zhu, and Li]{choi2025debate}
Hyeong~Kyu Choi, Xiaojin Zhu, and Yixuan Li.
\newblock Debate or vote: Which yields better decisions in multi-agent large language models?
\newblock \emph{arXiv preprint arXiv:2508.17536}, 2025.

\bibitem[Chrabaszcz et~al.(2017)Chrabaszcz, Loshchilov, and Hutter]{chrabaszcz2017downsampled}
Patryk Chrabaszcz, Ilya Loshchilov, and Frank Hutter.
\newblock A downsampled variant of imagenet as an alternative to the cifar datasets.
\newblock \emph{arXiv preprint arXiv:1707.08819}, 2017.

\bibitem[Cordts et~al.(2016)Cordts, Omran, Ramos, Rehfeld, Enzweiler, Benenson, Franke, Roth, and Schiele]{Cordts2016Cityscapes}
Marius Cordts, Mohamed Omran, Sebastian Ramos, Timo Rehfeld, Markus Enzweiler, Rodrigo Benenson, Uwe Franke, Stefan Roth, and Bernt Schiele.
\newblock The cityscapes dataset for semantic urban scene understanding.
\newblock In \emph{Proc. of the IEEE Conference on Computer Vision and Pattern Recognition (CVPR)}, 2016.

\bibitem[Dabney et~al.(2020)Dabney, Ostrovski, and Barreto]{dabney2020temporally}
Will Dabney, Georg Ostrovski, and Andr{\'e} Barreto.
\newblock Temporally-extended $\{$$\backslash$epsilon$\}$-greedy exploration.
\newblock \emph{arXiv preprint arXiv:2006.01782}, 2020.

\bibitem[Do{\u{g}}ru et~al.(2025)Do{\u{g}}ru, Keskin, and Aydo{\u{g}}an]{dougru2025taking}
An{\i}l Do{\u{g}}ru, Mehmet~Onur Keskin, and Reyhan Aydo{\u{g}}an.
\newblock Taking into account opponent’s arguments in human-agent negotiations.
\newblock \emph{ACM Transactions on Interactive Intelligent Systems}, 15\penalty0 (1):\penalty0 1--35, 2025.

\bibitem[Dong and Yang(2019{\natexlab{a}})]{dong2019one}
Xuanyi Dong and Yi Yang.
\newblock One-shot neural architecture search via self-evaluated template network.
\newblock In \emph{Proceedings of the IEEE/CVF international conference on computer vision}, pages 3681--3690, 2019{\natexlab{a}}.

\bibitem[Dong and Yang(2019{\natexlab{b}})]{dong2019searching}
Xuanyi Dong and Yi Yang.
\newblock Searching for a robust neural architecture in four gpu hours.
\newblock In \emph{Proceedings of the IEEE/CVF conference on computer vision and pattern recognition}, pages 1761--1770, 2019{\natexlab{b}}.

\bibitem[Dong and Yang(2020)]{dong2020bench}
Xuanyi Dong and Yi Yang.
\newblock Nas-bench-201: Extending the scope of reproducible neural architecture search.
\newblock \emph{arXiv preprint arXiv:2001.00326}, 2020.

\bibitem[Du et~al.(2023)Du, Li, Torralba, Tenenbaum, and Mordatch]{du2023improving}
Yilun Du, Shuang Li, Antonio Torralba, Joshua~B Tenenbaum, and Igor Mordatch.
\newblock Improving factuality and reasoning in language models through multiagent debate.
\newblock In \emph{Forty-first International Conference on Machine Learning}, 2023.

\bibitem[Falkner et~al.(2018)Falkner, Klein, and Hutter]{falkner2018bohb}
Stefan Falkner, Aaron Klein, and Frank Hutter.
\newblock Bohb: Robust and efficient hyperparameter optimization at scale.
\newblock In \emph{International conference on machine learning}, pages 1437--1446. PMLR, 2018.

\bibitem[Feng et~al.(2024)Feng, Su, Zheng, Ren, Zhang, Wu, Wang, and Liu]{feng2024m}
Zhaopeng Feng, Jiayuan Su, Jiamei Zheng, Jiahan Ren, Yan Zhang, Jian Wu, Hongwei Wang, and Zuozhu Liu.
\newblock M-mad: Multidimensional multi-agent debate for advanced machine translation evaluation.
\newblock \emph{arXiv preprint arXiv:2412.20127}, 2024.

\bibitem[Guo et~al.(2020)Guo, Zhang, Mu, Heng, Liu, Wei, and Sun]{guo2020single}
Zichao Guo, Xiangyu Zhang, Haoyuan Mu, Wen Heng, Zechun Liu, Yichen Wei, and Jian Sun.
\newblock Single path one-shot neural architecture search with uniform sampling.
\newblock In \emph{European conference on computer vision}, pages 544--560. Springer, 2020.

\bibitem[He et~al.(2016)He, Zhang, Ren, and Sun]{he2016deep}
Kaiming He, Xiangyu Zhang, Shaoqing Ren, and Jian Sun.
\newblock Deep residual learning for image recognition.
\newblock In \emph{Proceedings of the IEEE conference on computer vision and pattern recognition}, pages 770--778, 2016.

\bibitem[He et~al.(2025)He, Huang, Wang, Ran, Sheng, Huang, Lin, Xu, Liu, and Feng]{he2025crab}
Kai He, Yucheng Huang, Wenqing Wang, Delong Ran, Dongming Sheng, Junxuan Huang, Qika Lin, Jiaxing Xu, Wenqiang Liu, and Mengling Feng.
\newblock Crab: A novel configurable role-playing llm with assessing benchmark.
\newblock In \emph{Proceedings of the 63rd Annual Meeting of the Association for Computational Linguistics (Volume 1: Long Papers)}, pages 15030--15052, 2025.

\bibitem[Hu et~al.(2020)Hu, Xie, Zheng, Liu, Shi, Liu, and Lin]{hu2020dsnas}
Shoukang Hu, Sirui Xie, Hehui Zheng, Chunxiao Liu, Jianping Shi, Xunying Liu, and Dahua Lin.
\newblock Dsnas: Direct neural architecture search without parameter retraining.
\newblock In \emph{Proceedings of the IEEE/CVF conference on computer vision and pattern recognition}, pages 12084--12092, 2020.

\bibitem[Kaesberg et~al.(2025)Kaesberg, Becker, Wahle, Ruas, and Gipp]{kaesberg2025voting}
Lars~Benedikt Kaesberg, Jonas Becker, Jan~Philip Wahle, Terry Ruas, and Bela Gipp.
\newblock Voting or consensus? decision-making in multi-agent debate.
\newblock \emph{arXiv preprint arXiv:2502.19130}, 2025.

\bibitem[Kim et~al.(2024)Kim, Park, Jeong, Chan, Xu, McDuff, Lee, Ghassemi, Breazeal, and Park]{kim2024mdagents}
Yubin Kim, Chanwoo Park, Hyewon Jeong, Yik~S Chan, Xuhai Xu, Daniel McDuff, Hyeonhoon Lee, Marzyeh Ghassemi, Cynthia Breazeal, and Hae~W Park.
\newblock Mdagents: An adaptive collaboration of llms for medical decision-making.
\newblock \emph{Advances in Neural Information Processing Systems}, 37:\penalty0 79410--79452, 2024.

\bibitem[Kirillov et~al.(2019)Kirillov, Girshick, He, and Doll{\'a}r]{kirillov2019panoptic}
Alexander Kirillov, Ross Girshick, Kaiming He, and Piotr Doll{\'a}r.
\newblock Panoptic feature pyramid networks.
\newblock In \emph{Proceedings of the IEEE/CVF conference on computer vision and pattern recognition}, pages 6399--6408, 2019.

\bibitem[Krizhevsky et~al.(2009)Krizhevsky, Hinton, et~al.]{krizhevsky2009learning}
Alex Krizhevsky, Geoffrey Hinton, et~al.
\newblock Learning multiple layers of features from tiny images.(2009), 2009.

\bibitem[Li et~al.(2023)Li, Hammoud, Itani, Khizbullin, and Ghanem]{li2023camel}
Guohao Li, Hasan Hammoud, Hani Itani, Dmitrii Khizbullin, and Bernard Ghanem.
\newblock Camel: Communicative agents for" mind" exploration of large language model society.
\newblock \emph{Advances in Neural Information Processing Systems}, 36:\penalty0 51991--52008, 2023.

\bibitem[Li and Talwalkar(2020)]{li2020random}
Liam Li and Ameet Talwalkar.
\newblock Random search and reproducibility for neural architecture search.
\newblock In \emph{Uncertainty in artificial intelligence}, pages 367--377. PMLR, 2020.

\bibitem[Liang et~al.(2023)Liang, He, Jiao, Wang, Wang, Wang, Yang, Shi, and Tu]{liang2023encouraging}
Tian Liang, Zhiwei He, Wenxiang Jiao, Xing Wang, Yan Wang, Rui Wang, Yujiu Yang, Shuming Shi, and Zhaopeng Tu.
\newblock Encouraging divergent thinking in large language models through multi-agent debate.
\newblock \emph{arXiv preprint arXiv:2305.19118}, 2023.

\bibitem[Lin et~al.(2014)Lin, Maire, Belongie, Hays, Perona, Ramanan, Doll{\'a}r, and Zitnick]{lin2014microsoft}
Tsung-Yi Lin, Michael Maire, Serge Belongie, James Hays, Pietro Perona, Deva Ramanan, Piotr Doll{\'a}r, and C~Lawrence Zitnick.
\newblock Microsoft coco: Common objects in context.
\newblock In \emph{European conference on computer vision}, pages 740--755. Springer, 2014.

\bibitem[Liu et~al.(2024)Liu, Feng, Xue, Wang, Wu, Lu, Zhao, Deng, Zhang, Ruan, et~al.]{liu2024deepseek}
Aixin Liu, Bei Feng, Bing Xue, Bingxuan Wang, Bochao Wu, Chengda Lu, Chenggang Zhao, Chengqi Deng, Chenyu Zhang, Chong Ruan, et~al.
\newblock Deepseek-v3 technical report.
\newblock \emph{arXiv preprint arXiv:2412.19437}, 2024.

\bibitem[Liu et~al.(2018)Liu, Simonyan, and Yang]{liu2018darts}
Hanxiao Liu, Karen Simonyan, and Yiming Yang.
\newblock Darts: Differentiable architecture search.
\newblock \emph{arXiv preprint arXiv:1806.09055}, 2018.

\bibitem[Liu et~al.(2019)Liu, Simonyan, and Yang]{liu2019dartsdifferentiablearchitecturesearch}
Hanxiao Liu, Karen Simonyan, and Yiming Yang.
\newblock Darts: Differentiable architecture search, 2019.

\bibitem[Liu et~al.(2025)Liu, Shi, Song, and Xu]{liu2025dual}
Yutong Liu, Lida Shi, Rui Song, and Hao Xu.
\newblock A dual-mind framework for strategic and expressive negotiation agent.
\newblock In \emph{Proceedings of the 63rd Annual Meeting of the Association for Computational Linguistics (Volume 1: Long Papers)}, pages 23840--23860, 2025.

\bibitem[Movahedi et~al.(2022)Movahedi, Adabinejad, Imani, Keshavarz, Dehghani, Shakery, and Araabi]{movahedi2022lambda}
Sajad Movahedi, Melika Adabinejad, Ayyoob Imani, Arezou Keshavarz, Mostafa Dehghani, Azadeh Shakery, and Babak~N Araabi.
\newblock Lambda-darts: Mitigating performance collapse by harmonizing operation selection among cells.
\newblock \emph{arXiv preprint arXiv:2210.07998}, 2022.

\bibitem[Nan et~al.(2023)Nan, Zhang, Zou, Zhao, Zhou, and Cohan]{nan2023evaluating}
Linyong Nan, Ellen Zhang, Weijin Zou, Yilun Zhao, Wenfei Zhou, and Arman Cohan.
\newblock On evaluating the integration of reasoning and action in llm agents with database question answering.
\newblock \emph{arXiv preprint arXiv:2311.09721}, 2023.

\bibitem[Nasir et~al.(2024)Nasir, Earle, Togelius, James, and Cleghorn]{nasir2024llmatic}
Muhammad~Umair Nasir, Sam Earle, Julian Togelius, Steven James, and Christopher Cleghorn.
\newblock Llmatic: neural architecture search via large language models and quality diversity optimization.
\newblock In \emph{proceedings of the Genetic and Evolutionary Computation Conference}, pages 1110--1118, 2024.

\bibitem[Nguyen et~al.(2021)Nguyen, Le, Yamada, and Osborne]{nguyen2021optimal}
Vu Nguyen, Tam Le, Makoto Yamada, and Michael~A Osborne.
\newblock Optimal transport kernels for sequential and parallel neural architecture search.
\newblock In \emph{International Conference on Machine Learning}, pages 8084--8095. PMLR, 2021.

\bibitem[Park et~al.(2024)Park, Kim, Jin, Park, and Han]{park2024predict}
Someen Park, Jaehoon Kim, Seungwan Jin, Sohyun Park, and Kyungsik Han.
\newblock Predict: multi-agent-based debate simulation for generalized hate speech detection.
\newblock In \emph{Proceedings of the 2024 Conference on Empirical Methods in Natural Language Processing}, pages 20963--20987, 2024.

\bibitem[Pham et~al.(2018)Pham, Guan, Zoph, Le, and Dean]{pham2018efficient}
Hieu Pham, Melody Guan, Barret Zoph, Quoc Le, and Jeff Dean.
\newblock Efficient neural architecture search via parameters sharing.
\newblock In \emph{International conference on machine learning}, pages 4095--4104. PMLR, 2018.

\bibitem[Rahman and Chakraborty(2024)]{rahman2024lemo}
Md~Hafizur Rahman and Prabuddha Chakraborty.
\newblock Lemo-nade: Multi-parameter neural architecture discovery with llms.
\newblock \emph{arXiv preprint arXiv:2402.18443}, 2024.

\bibitem[Real et~al.(2019)Real, Aggarwal, Huang, and Le]{real2019regularized}
Esteban Real, Alok Aggarwal, Yanping Huang, and Quoc~V Le.
\newblock Regularized evolution for image classifier architecture search.
\newblock In \emph{Proceedings of the aaai conference on artificial intelligence}, pages 4780--4789, 2019.

\bibitem[Rodrigues~Gomes and Kowalczyk(2009)]{rodrigues2009dynamic}
Eduardo Rodrigues~Gomes and Ryszard Kowalczyk.
\newblock Dynamic analysis of multiagent q-learning with $\varepsilon$-greedy exploration.
\newblock In \emph{Proceedings of the 26th annual international conference on machine learning}, pages 369--376, 2009.

\bibitem[Shen et~al.(2023)Shen, Song, Tan, Li, Lu, and Zhuang]{shen2023hugginggpt}
Yongliang Shen, Kaitao Song, Xu Tan, Dongsheng Li, Weiming Lu, and Yueting Zhuang.
\newblock Hugginggpt: Solving ai tasks with chatgpt and its friends in hugging face, 2023.
\newblock \emph{arXiv preprint arXiv:2303.17580}, 2023.

\bibitem[Shinn et~al.(2023)Shinn, Cassano, Gopinath, Narasimhan, and Yao]{shinn2023reflexion}
Noah Shinn, Federico Cassano, Ashwin Gopinath, Karthik Narasimhan, and Shunyu Yao.
\newblock Reflexion: Language agents with verbal reinforcement learning.
\newblock \emph{Advances in Neural Information Processing Systems}, 36:\penalty0 8634--8652, 2023.

\bibitem[Tian et~al.(2020)Tian, Wang, Huang, Li, Dai, Yang, Wang, and Fink]{tian2020off}
Yuan Tian, Qin Wang, Zhiwu Huang, Wen Li, Dengxin Dai, Minghao Yang, Jun Wang, and Olga Fink.
\newblock Off-policy reinforcement learning for efficient and effective gan architecture search.
\newblock In \emph{European conference on computer vision}, pages 175--192. Springer, 2020.

\bibitem[Tokic(2010)]{tokic2010adaptive}
Michel Tokic.
\newblock Adaptive $\varepsilon$-greedy exploration in reinforcement learning based on value differences.
\newblock In \emph{Annual conference on artificial intelligence}, pages 203--210. Springer, 2010.

\bibitem[Wang et~al.(2022)Wang, Wei, Schuurmans, Le, Chi, Narang, Chowdhery, and Zhou]{wang2022self}
Xuezhi Wang, Jason Wei, Dale Schuurmans, Quoc Le, Ed Chi, Sharan Narang, Aakanksha Chowdhery, and Denny Zhou.
\newblock Self-consistency improves chain of thought reasoning in language models.
\newblock \emph{arXiv preprint arXiv:2203.11171}, 2022.

\bibitem[Wang et~al.(2025)Wang, Zhang, Zheng, Ai, and Wang]{wang2025debt}
Xiaofeng Wang, Zhixin Zhang, Jinguang Zheng, Yiming Ai, and Rui Wang.
\newblock Debt collection negotiations with large language models: An evaluation system and optimizing decision making with multi-agent.
\newblock \emph{arXiv preprint arXiv:2502.18228}, 2025.

\bibitem[Williams(1992)]{williams1992simple}
Ronald~J Williams.
\newblock Simple statistical gradient-following algorithms for connectionist reinforcement learning.
\newblock \emph{Machine learning}, 8\penalty0 (3):\penalty0 229--256, 1992.

\bibitem[Wu et~al.(2023)Wu, Yin, Qi, Wang, Tang, and Duan]{wu2023visual}
Chenfei Wu, Shengming Yin, Weizhen Qi, Xiaodong Wang, Zecheng Tang, and Nan Duan.
\newblock Visual chatgpt: Talking, drawing and editing with visual foundation models.
\newblock \emph{arXiv preprint arXiv:2303.04671}, 2023.

\bibitem[Wu et~al.(2024)Wu, Bansal, Zhang, Wu, Li, Zhu, Jiang, Zhang, Zhang, Liu, et~al.]{wu2024autogen}
Qingyun Wu, Gagan Bansal, Jieyu Zhang, Yiran Wu, Beibin Li, Erkang Zhu, Li Jiang, Xiaoyun Zhang, Shaokun Zhang, Jiale Liu, et~al.
\newblock Autogen: Enabling next-gen llm applications via multi-agent conversations.
\newblock In \emph{First Conference on Language Modeling}, 2024.

\bibitem[Xi et~al.(2025)Xi, Chen, Guo, He, Ding, Hong, Zhang, Wang, Jin, Zhou, et~al.]{xi2025rise}
Zhiheng Xi, Wenxiang Chen, Xin Guo, Wei He, Yiwen Ding, Boyang Hong, Ming Zhang, Junzhe Wang, Senjie Jin, Enyu Zhou, et~al.
\newblock The rise and potential of large language model based agents: A survey.
\newblock \emph{Science China Information Sciences}, 68\penalty0 (2):\penalty0 121101, 2025.

\bibitem[Xie et~al.(2018)Xie, Zheng, Liu, and Lin]{xie2018snas}
Sirui Xie, Hehui Zheng, Chunxiao Liu, and Liang Lin.
\newblock Snas: stochastic neural architecture search.
\newblock \emph{arXiv preprint arXiv:1812.09926}, 2018.

\bibitem[Xu et~al.(2019)Xu, Xie, Zhang, Chen, Qi, Tian, and Xiong]{xu2019pc}
Yuhui Xu, Lingxi Xie, Xiaopeng Zhang, Xin Chen, Guo-Jun Qi, Qi Tian, and Hongkai Xiong.
\newblock Pc-darts: Partial channel connections for memory-efficient architecture search.
\newblock \emph{arXiv preprint arXiv:1907.05737}, 2019.

\bibitem[Yang et~al.(2025)Yang, Zeng, Jin, Qian, Luo, and Liu]{yang2025nader}
Zekang Yang, Wang Zeng, Sheng Jin, Chen Qian, Ping Luo, and Wentao Liu.
\newblock Nader: Neural architecture design via multi-agent collaboration.
\newblock In \emph{Proceedings of the Computer Vision and Pattern Recognition Conference}, pages 4452--4461, 2025.

\bibitem[Ye et~al.(2022)Ye, Li, Li, Chen, Fan, and Ouyang]{ye2022b}
Peng Ye, Baopu Li, Yikang Li, Tao Chen, Jiayuan Fan, and Wanli Ouyang.
\newblock b-darts: Beta-decay regularization for differentiable architecture search.
\newblock In \emph{proceedings of the IEEE/CVF conference on computer vision and pattern recognition}, pages 10874--10883, 2022.

\bibitem[Yu et~al.(2024)Yu, Yu, Wei, Zhang, and Qian]{yu2024beyond}
Yeyong Yu, Runsheng Yu, Haojie Wei, Zhanqiu Zhang, and Quan Qian.
\newblock Beyond dialogue: A profile-dialogue alignment framework towards general role-playing language model.
\newblock \emph{arXiv preprint arXiv:2408.10903}, 2024.

\bibitem[Zhang et~al.(2025{\natexlab{a}})Zhang, Luo, Liu, Wu, Lin, Zeng, Qu, Fang, Yang, Gao, et~al.]{zhang2025omnicharacter}
Haonan Zhang, Run Luo, Xiong Liu, Yuchuan Wu, Ting-En Lin, Pengpeng Zeng, Qiang Qu, Feiteng Fang, Min Yang, Lianli Gao, et~al.
\newblock Omnicharacter: Towards immersive role-playing agents with seamless speech-language personality interaction.
\newblock \emph{arXiv preprint arXiv:2505.20277}, 2025{\natexlab{a}}.

\bibitem[Zhang et~al.(2021)Zhang, Su, Pan, Chang, Abbasnejad, and Haffari]{zhang2021idarts}
Miao Zhang, Steven~W Su, Shirui Pan, Xiaojun Chang, Ehsan~M Abbasnejad, and Reza Haffari.
\newblock idarts: Differentiable architecture search with stochastic implicit gradients.
\newblock In \emph{International Conference on Machine Learning}, pages 12557--12566. PMLR, 2021.

\bibitem[Zhang et~al.(2025{\natexlab{b}})Zhang, An, Qiao, Yu, Chen, Wang, Yin, Sun, and Zhang]{zhang2025roleplot}
Pinyi Zhang, Siyu An, Lingfeng Qiao, Yifei Yu, Jingyang Chen, Jie Wang, Di Yin, Xing Sun, and Kai Zhang.
\newblock Roleplot: A systematic framework for evaluating and enhancing the plot-progression capabilities of role-playing agents.
\newblock In \emph{Proceedings of the 63rd Annual Meeting of the Association for Computational Linguistics (Volume 1: Long Papers)}, pages 12337--12354, 2025{\natexlab{b}}.

\bibitem[Zhao et~al.(2024{\natexlab{a}})Zhao, Huang, Xu, Lin, Liu, and Huang]{zhao2024expel}
Andrew Zhao, Daniel Huang, Quentin Xu, Matthieu Lin, Yong-Jin Liu, and Gao Huang.
\newblock Expel: Llm agents are experiential learners.
\newblock In \emph{Proceedings of the AAAI Conference on Artificial Intelligence}, pages 19632--19642, 2024{\natexlab{a}}.

\bibitem[Zhao et~al.(2024{\natexlab{b}})Zhao, Zhang, Chia, Xu, Zhao, and Bing]{zhao2024auto}
Ruochen Zhao, Wenxuan Zhang, Yew~Ken Chia, Weiwen Xu, Deli Zhao, and Lidong Bing.
\newblock Auto-arena: Automating llm evaluations with agent peer battles and committee discussions.
\newblock \emph{arXiv preprint arXiv:2405.20267}, 2024{\natexlab{b}}.

\bibitem[Zheng et~al.(2023)Zheng, Su, You, Wang, Qian, Xu, and Albanie]{zheng2023can}
Mingkai Zheng, Xiu Su, Shan You, Fei Wang, Chen Qian, Chang Xu, and Samuel Albanie.
\newblock Can gpt-4 perform neural architecture search?
\newblock \emph{arXiv preprint arXiv:2304.10970}, 2023.

\end{thebibliography}
